\documentclass{article}
\usepackage{ijcai22}

\usepackage{times}

\usepackage{soul}
\usepackage{url}
\usepackage{anyfontsize}
\usepackage[11pt]{moresize}
\usepackage[hidelinks]{hyperref}
\usepackage[utf8]{inputenc}
\usepackage[small]{caption}
\usepackage{graphicx}
\usepackage{amsmath}
\usepackage{booktabs}
\usepackage{hyperref}
\usepackage{url}
\usepackage{graphicx}
\usepackage{amsmath}
\usepackage{booktabs}
\usepackage{color}
\usepackage{latexsym}
\usepackage{bm}
\usepackage{algorithm}
\usepackage{algorithmic}
\usepackage{subfigure}
\usepackage{booktabs}
\usepackage{xcolor}
\usepackage{array}
\usepackage{amssymb}
\usepackage{wrapfig}
\usepackage{xr}
\usepackage{mathptmx}
\usepackage{anyfontsize}
\usepackage{t1enc}

\usepackage{xspace} % Needed for et al.
\newcommand{\ie}{\emph{i.e.,}\xspace}
\newcommand{\eg}{\emph{e.g.,}\xspace}

\title{Spiking Graph Convolutional Networks}

\author{
Zulun Zhu$^{1,2}$\footnote{This work was done when the first author was with Rochester Institute of Technology.}\and
Jiaying Peng$^1$\and
Jintang Li $^1$\and
Liang Chen$^1$\footnote{Corresponding Author}\and
Qi Yu $^2$\And
Siqiang Luo $^3$
\affiliations
$^1$Sun Yat-Sen University\\
$^2$Rochester Institute of Technology\\
$^3$Nanyang Technological University
\emails
zulun.zhu@gmail.com,
\{pengjy36,lijt55\}@mail2.sysu.edu.cn,
chenliang6@mail.sysu.edu.cn,
qi.yu@rit.edu,
siqiang.luo@ntu.edu.sg
}

\begin{document}

\maketitle

\begin{abstract}
Graph Convolutional Networks (GCNs) achieve an impressive performance due to the remarkable representation ability in learning the graph information. However, GCNs, when implemented on a deep network, require expensive computation power, making them difficult to be deployed on battery-powered devices. In contrast, Spiking Neural Networks (SNNs), which perform a bio-fidelity inference process, offer an energy-efficient neural architecture. In this work, we propose SpikingGCN, an end-to-end framework that aims to integrate the embedding of GCNs with the biofidelity characteristics of SNNs. The original graph data are encoded into spike trains based on the incorporation of graph convolution. We further model biological information processing
by utilizing a fully connected layer combined with neuron nodes. In a wide range of scenarios (\eg citation networks, image graph classification, and recommender systems), our experimental results show that the proposed method could gain competitive performance against state-of-the-art approaches. Furthermore, we show that SpikingGCN on a neuromorphic chip can bring a clear advantage of energy efficiency into graph data analysis, which demonstrates its great potential to construct environment-friendly machine learning models.
\end{abstract}
\section{Introduction}

Graph Neural Networks (GNNs), especially those using convolutional methods, have become a popular computational model for graph data analysis as the high-performance computing systems blossom during the last decade.
One of well-known methods in GNNs is Graph Convolutional Networks (GCNs) \cite{kipf2016semi}, which learn a high-order approximation of a spectral graph by using convolutional layers followed by a nonlinear activation function to make the final prediction. Like most of the deep learning models, GCNs incorporate complex structures with costly training and testing process, leading to significant power consumption. It has been reported that the computation resources consumed for deep learning have grown $300,000$-fold from 2012 to 2018 \cite{DBLP:journals/corr/abs-2007-03051}.
%which could hardly be a clear trend to need-driven complexity.
The high energy consumption, when further coupled with sophisticated theoretical analysis and blurred biological interpretability of the network, has resulted in a revival of effort in developing novel energy-efficient neural architectures and physical hardware.

Inspired by the brain-like computing process, Spiking Neural Networks (SNNs) formalize the event- or clock-driven signals as inference for a set of parameters to update the neuron nodes \cite{brette2007simulation}. 
 Different from conventional deep learning models that communicate information using continuous decimal values, SNNs perform inexpensive computation by transmitting the input into discrete spike trains. 
 Such a bio-fidelity method can perform a more intuitive and simpler inference and model training than traditional networks \cite{maass1997networks,zhang2022recent}. Another distinctive merit of SNNs is the intrinsic power efficiency on the neuromorphic hardware, which is capable of running 1 million neurons and 256 million synapses with only 70 mW energy cost \cite{merolla2014million}. Nevertheless, employing SNNs as an energy-efficient architecture to process graph data as effectively as GCNs still faces fundamental challenges.

\paragraph{Challenges.} (i) \emph{Spike representation.} Despite the promising results achieved  on common tasks (\eg image classification), SNN models 
%the achievement for the image classification problem at the current stage, most SNN models 
are not trivially portable to non-Euclidean domains, such as graphs. Given the graph datasets widely used in many applications (\eg citation networks and social networks), how to extract the graph structure and transfer the graph data into spike trains poses a challenge. (ii) \emph{Model generalization.} GCNs can be extended to diverse circumstances by using deeper layers. Thus, it is essential to further extend the SNNs to a wider scope of applications where graphs are applicable. (iii) \emph{Energy efficiency.} Except for the common metrics like accuracy or 
prediction loss in artificial neural networks (ANNs), the energy efficiency of SNNs on the neuromorphic chips is an important characteristic to be considered. However, neuromorphic chips are not as advanced as contemporary GPUs, and the lack of uniform standards also impacts the energy estimation on different platforms.

To tackle these fundamental challenges, we introduce Spiking Graph Neural Network (SpikingGCN): an end-to-end framework that can properly encode graphs and make a prediction for non-trivial graph datasets that arise in diverse domains. To our best knowledge, {\em SpikingGCN is the first-ever SNN designed for node classification in graph data}, and it can also be extended into more complex neural network structures. Overall, our main contribution is threefold: (i) We propose SpikingGCN, the first end-to-end model for node classification in SNNs, without any pre-training and conversion. The graph data is transformed into spike trains by a spike encoder. These generated spikes are used to predict the classification results. (ii) We show that the basic model inspired by GCNs can effectively merge the convolutional features into spikes and achieve competitive predictive performance. In addition, we further evaluate the performance of our model for active learning and energy efficient settings;
 (iii) We extend our framework to enable more complex network structures for different tasks,  including image graph classification and rating predictions in recommender systems. The extensibility of the proposed model also opens the gate to perform SNN-based inference and training in various kinds of graph-based data. The code and Appendix are available on Github\footnote{https://github.com/ZulunZhu/SpikingGCN.git}.

\section{Spiking Graph Neural Networks}
Graphs are usually represented by a non-Euclidean data structure consisting of a set of nodes (vertices) and their relationships (edges). The reasoning process in the human brain depends heavily on the graph extracted from daily experience \cite{DBLP:journals/aiopen/ZhouCHZYLWLS20}. However, how to perform biologically interpretable reasoning for the standard graph neural networks has not been adequately investigated. %We provide the limitation of other related work in Appendix \ref{related work}. 
Thus, the proposed SpikingGCN aims to address challenges of semi-supervised node classification in a biological and energy-efficient fashion. As this work refers to the methods in GNNs and SNNs, we list the frequently used notations in Table 8 in Appendix.

Graph neural networks (GNNs) conduct propagation guided by the graph structure, which is fundamentally different from existing SNN models that can only handle relatively simple image data. Instead of treating the single node as the input of an SNN model, the states of their neighborhood should also be considered. Let $\mathcal{G}=(\mathcal{V},\bm{A})$ formally denote a graph, where $\mathcal{V}$ is the node set $\{v_1,...,v_N\}$ and $\bm{A}\in\mathbb{R}^{N\times{N}}$ represents the adjacent matrix. Here $N$ is the number of nodes. The entire attribute matrix $\bm{X}\in\mathbb{R}^{N\times{d}}$ includes the vectors of all nodes $[x_1,...,x_N]^\top$.
The degree matrix $\bm{D} = \text{diag}(d_1,...,d_N)$ consists of the row-sum of the adjacent matrix $d_i = \sum_j a_{ij}$, where $a_{ij}$ denotes the edge weight between nodes $v_i$ and $v_j$. Each node has $d$ dimensions. 
Our goal is to conduct SNN inference without neglecting the relationships between nodes.

 Inference in SNN models is commonly conducted through the classic Leaky Integrate-and-Fire (LIF) mechanism \cite{gerstner2002spiking}. Given the membrane potential $V_m^{t}$ at time step $t$, the time constant $\tau_m$, and the new pre-synaptic input $\Delta V_m$, the membrane potential activity is governed by:
\begin{equation}\label{equ:LIF}
    \tau_m\frac{dV_m^{t}}{dt}=-(V_m^{t}-V_{reset})+\Delta V_m,
\end{equation}
where $V_{reset}$ is the signed reset voltage. The left differential item is widely used in the continuous domain, but the biological simulation in SNNs requires the implementation to be executed in a discrete and sequential way. Thus, we approximate the differential expression using an iterative version to guarantee computational availability. Updating $\Delta V_m$ using the input $I(t)$ of our network, we can formalize \eqref{equ:LIF} as:
\begin{equation} \label{equ:charge_1}
    V_m^{t}=V_m^{t-1}+\frac{1}{\tau_m}(-V_m^{t-1}+V_{reset}+I(t)).
\end{equation}

\begin{figure}[ht]
\centering
\includegraphics[width=3.3in]{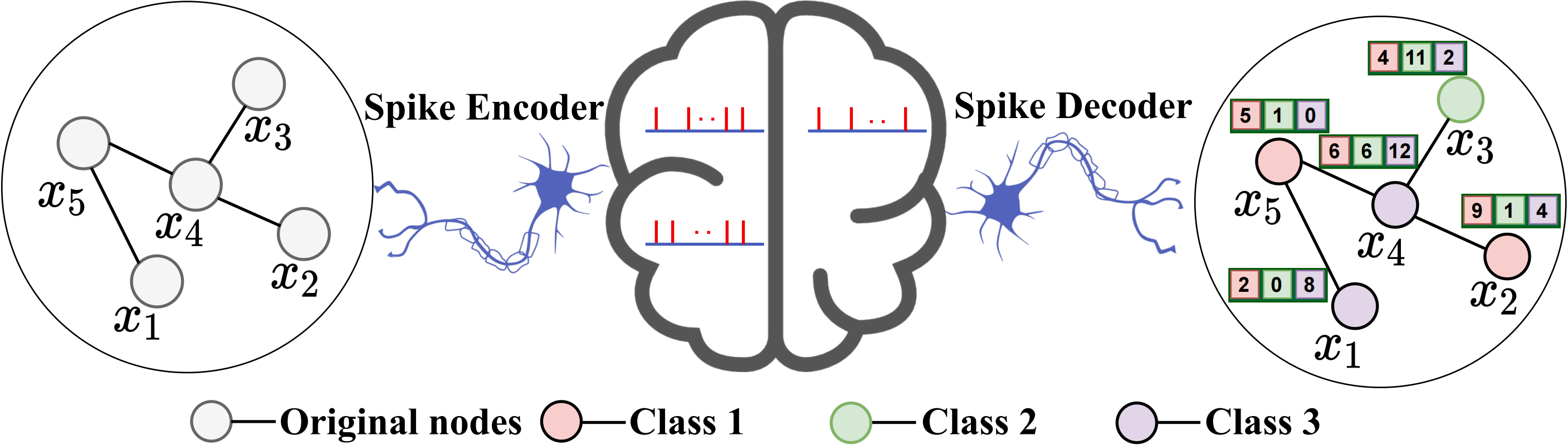} 
\caption{Schematic view of the proposed SpikingGCN. 
% The original graph nodes will be transformed into the spike trains for the SNN model via a spike encoder. The encoder should aggregate the attributes in the graph and represent them in the format of spikes. Finally, a spike decoder, which is always made up of linear layers, may convert spike trains into various firing rates. In this case, a greater firing rate implies a higher chance that a node belongs to one of the classes.  
}
\label{fig:overall}
\end{figure}

To tackle the issue of feature propagation in an SNN model, we consider a spike encoder to extract the information in the graph and output the hidden state of each node in the format of spike trains. As shown in Fig. \ref{fig:overall}, the original input graph is transformed into the spikes from a convolution perspective. To predict the labels for each node, we consider a spike decoder and treat the final spike rate as a classification result. %In Section \ref{sec:convolution}, we will explain the training and inference procedure in detail, denoted as the basic SpikingGCN model.

%\section{GNN meets SNN\label{sec:convolution}}

\paragraph{Graph Convolution.} 
The pattern of graph data consists of two parts: topological structure and node's own features, which are stored in the adjacency and attribute matrices, respectively. Different from the general processing of images with single-channel pixel features, the topological structure will be absent if only the node attributes are considered. To avoid the performance degradation of attributes-only encoding, SpikingGCN utilizes the graph convolution method inspired by GCNs to incorporate the topological information. The idea is to use the adjacency relationship to normalize the weights, thus nodes can selectively aggregate neighbor attributes. The convolution result, \ie node representations, will serve as input to the subsequent spike encoder. Following the propagation mechanism of GCN \cite{kipf2016semi} and SGC \cite{wu2019simplifying}, we form the new node representation $h_i$ utilizing the attributes $x_i$ of each node $v_i$ and its local neighborhood:
\begin{equation} \label{equ:conv1}
    h_i\gets \frac{1}{d_i+1}x_i+\sum_{j=1}^N\frac{a_{ij}}{\sqrt{(d_i+1)(d_j+1)}}x_j.
\end{equation}
Here, we can express the attribute transformation over the entire graph by:
\begin{equation} \label{equ:conv2}
   \bm{S} = \Tilde{\bm{D}}^{-\frac{1}{2}}\Tilde{\bm{A}}\Tilde{\bm{D}}^{-\frac{1}{2}} ,\bm{H} = \bm{S}^K \bm{X},
\end{equation}

where $\Tilde{\bm{A}}=\bm{A}+\bm{I}$ is the adjacent matrix with added self-connection, $K$ is the graph convolution layer number and $\bm{\Tilde{D}}$ is the degree matrix of $\Tilde{\bm{A}}$. Similar to the simplified framework as SGC, we drop the non-linear operation and focus on the convolutional process on the entire graph. As a result, \eqref{equ:conv2} acts as the only convolution operation in the spike encoder. While we incorporate the feature propagation explored by GCN and SGC, we would like to further highlight our novel contributions. First, our original motivation is to leverage an SNNs-based framework to reduce the inference energy consumption of graph analysis tasks without performance degradation. GCN's effective graph Laplacian regularization approach allows us to minimize the number of trainable parameters and perform efficient inference in SNNs. Second, convolutional techniques only serve as the initial building block of SpikingGCN. More significantly, SpikingGCN is designed to accept the convolutional results in a binary form (spikes), and  further detect the specific patterns among these spikes. This biological mechanism makes it suitable to be deployed on a neuromorphic chip to improve energy efficiency.    

\paragraph{Representation Encoding.}
The representation $\bm{H}$ consists of continuous float-point values, but SNNs accept discrete spike signals. A spike encoder is essential to take node representations as input and output spikes for the subsequent procedures. We propose to use a probability-based Bernoulli encoding scheme as the basic method to transform the node representations to the spike signals. Let $O^{pre}_{i,t} = (o_{i1},...,o_{id})$ and $\lambda_{ij}$ denote the spikes before the fully connected layers' neurons at the $t$-th time step and the $j$-th feature in the new representation for node $i$, respectively. Our hypothesis is that the spiking rate should keep a positive relationship with the importance of patterns in the representations. In probability-based Bernoulli encoder, the probability $p$ to fire a spike $o_{ij}$ by each feature is related to the value of $\lambda_{ij}$ in node representation as following: 
\begin{equation}\label{encoding}
    p(o_{ij}) \sim \text{Bernoulli}(\lambda_{ij}), \lambda_{ij}=\min(\lambda_{ij},1.0).
\end{equation}
Here, $o_{ij}$ with $j\in [d]$ denotes a pre-synaptic spike, which takes a binary value (0 or 1). Note that $\lambda_{ij}$ derived from the convolution of neighbors is positively correlated with the feature significance. The larger the value, the greater the chance of a spike being fired by the encoder. 
Since the encoder generates the spike for each node on a tiny scale, we interpret the encoding module as a sampling process of the entire graph. In order to fully describe the information in the graph, we use $T$ time steps to repeat the sampling process. It is noteworthy that the number of time steps can be defined as the resolution of the message encoded. 

\paragraph{Charge, Fire and Reset in SpikingGCN.}
The following module includes the fully connected layer and the LIF neuron layer. The fully connected layer takes spikes as input and outputs voltages according to trainable weights. The voltages charge LIF neurons and then conduct a series of actions, including fire spikes and reset the membrane potential. %In this section, we detail these neuronal actions: charge, fire, and reset.

\begin{figure*}[tbp]
\centering\includegraphics[width=6.5in]{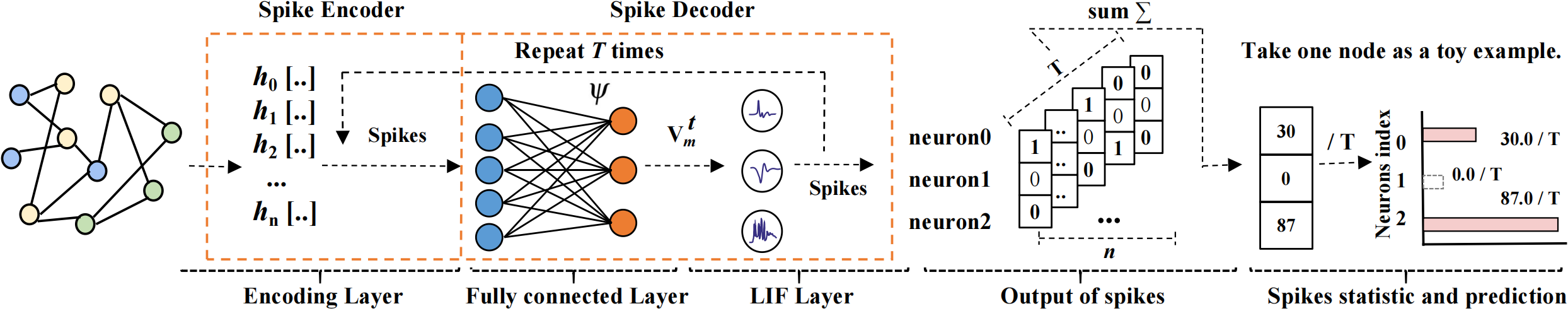}
\caption{An illustration of SpikingGCN's detailed framework} %We mark the spike encoder and spike decoder using the dotted box.}
\label{fig:framework}
\end{figure*}

{\em Potential charge.} General deep SNN models adopt a multi-layer network structure including linear and nonlinear counterparts to process the input \cite{cao2015spiking}. Following SGC's assumption, the depth in deep SNNs is not critical to predict unknown labels on the graph \cite{wu2019simplifying}. Thus, we drop redundant modules except for the final linear layer (fully connected layer) to simplify our framework and increase the inference speed. We obtain the linear summation $\sum_{j}\psi_j o_{ij}$ as the input of SNN structure in \eqref{equ:charge_1}. The LIF model includes the floating-point multiplication of the constant $\tau_{m}$, which is not biologically plausible. To address this challenge and avoid the additional hardware requirement when deployed on neuromorphic chips, we calculate the factor $1-\frac{1}{\tau_m}$ as $k_m$ and incorporate the constant $\frac{1}{\tau_m}$ into the synapse parameters $\psi$, then simplify the equation as:
\begin{equation} \label{equ:charge_2}
    V_m^{t}=k_mV_m^{t-1}+\sum_{j}\psi_j o_{ij}.
\end{equation}
{\em Fire and reset.} In a biological neuron, a spike is fired when the accumulated membrane potential passes a spiking threshold $V_{th}$. In essence, the spikes $O^{post}_{i,t}$ after the LIF neurons are generated and increase the spike rate in the output layer. We adopt the Heaviside function:
\begin{equation}\label{equ:heaviside}
\text {H}\left(V_m^{t}\right)=\left\{
\begin{array}{ll}
1 & \text { if } V_m^{t}\geq V_{\text {th}} \\
0 & \text { otherwise},
\end{array}\right.
\end{equation}
to simulate the fundamental firing process. As shown in Fig. \ref{fig:framework}, which demonstrates our framework, $T$ time spike trains for each node are generated from the three LIF neurons. Neurons sum the number of spikes and then divide it by $T$ to get the firing rate of each individual.
For instance, for the example nodes from ACM datasets, we get neurons' firing rates as: $fr=[30.0/T,0.0/T,87.0/T]^\top$, the true label: $lb=[0,0,1]^\top$, then the loss in training process is MSE($fr$, $lb$). If it is in the testing phase, the predicted label would be the  neuron with the max firing rate, \ie 2 in this example.

Negative voltages would not trigger spikes, but these voltages contain information that \eqref{equ:heaviside} ignores. To compensate for the negative term, we propose to use a negative threshold to distinguish the negative characteristics of the membrane potential. Inspired by \cite{kim2020spiking}, we adjust the Heaviside activation function after the neuron nodes as follows:
\begin{equation} \label{equ:new_heaviside}
\text {H}\left(V_m^{t}\right)=\left\{
\begin{array}{ll}
1 & \text { if } V_m^{t}\geq V_{\text {th }}, \\
-1 & \text { if } V_m^{t} \leq -\frac{1}{\theta} V_{\text {th }}, \\
0 & \text { otherwise,}
\end{array}\right.
\end{equation}
where $\theta$ is the hyperparameter that determines the negative range. In the context of biological mechanisms, we interpret the fixed activation function as an excitatory and inhibitory processes in the neurons. When capturing more information and firing spikes in response to various features, this more biologically reasonable modification also improves the performance of our model on classification tasks. The whole process is detailed by Algorithm 1 in Appendix B.

In biological neural systems, after firing a spike, the neurons tend to rest their potential and start to accumulate voltage again.
We reset the membrane potential: 
\begin{equation}\label{equ:reset}
    V^t_m=V^t_m-V_{th}.
\end{equation}

% \paragraph{Gradient Surrogate.}  One of the most significant obstacles for SNN's training is the non-differentiable nature of activation function \eqref{equ:heaviside}. In that case, the back-propagation algorithm can not be employed directly during the training phase. Inspired by \cite{roy2019scaling}, the sigmoid function is adopted to approximate the training fire phase when executing the back-propagation and stochastic gradient descent operation. Thus, we formalize the surrogate function as follow:
% \begin{equation} \label{equ:surragate}
%     H(V^t_m)\approx G(V^t_m)=\frac{1}{1+\exp(-\alpha V^t_m))},
% \end{equation}
% where $\alpha$ can measure the approximation degree. When we train the model, the back-propagation is feasible and the gradient of the Heaviside function can be replaced as followed:
% \begin{equation}
% \frac{\partial H(V^t_m)}{\partial V^t_m} \approx \frac{\partial G(V^t_m)}{\partial V^t_m}=\frac{\alpha\cdot \exp(-\alpha V^t_m)}{(1+\exp(-\alpha V^t_m))^2}.
% \end{equation}

\paragraph{Model Feasibility Analysis.}
Since the input spikes can be viewed as a rough approximation of original convolutional results in the initial graph,  two key questions remain: (i) does this proposed method really work for the prediction task? (ii) how to control the information reduction of the sampled spikes compared with other GNN models? It turns out that although equation \eqref{equ:charge_2} shows a rough intuitive approximation of the graph input using trainable linear combination, it cannot fully explain why SpikingGCN can achieve comparable performance with other real-value GNN models. Next, we will show that our spike representations $\psi_j o_j$ is very close to the real-value output with a high probability.

To explain why SpikingGCN can provide an accurate output using spike trains, let us list SGC (\cite{wu2019simplifying}) as the model for comparison. SGC adopts a similar framework to our model, which transfers the convolution result $\bm{H}$ into a fully connected layer in a real-value format. Given the parameters $\psi = (\psi_{1}, ...\psi_{j},..., \psi_{d} )$, the real-value convolution result $h = (\lambda_1, ..., \lambda_j, ..., \lambda_d)$ and the spike representations $O^{pre}_{i,t} = (o_{1},...,o_j,...,o_{d})$ for one node, we have:
\begin{equation}\label{equ:approximation}
    \operatorname{Pr}\left(o_{j}=1\right)=\lambda_{j}, \quad \operatorname{Pr}\left(o_{j}=0\right)=1-\lambda_{j}
\end{equation}
Note that firing a spike in each dimension of a node is independent. When merging our case into a generalization of the Chernoff inequalities for binomial distribution in \cite{chung2002connected}, we  derive the following estimation error bound.

Let $o_{1},...,o_j,...,o_{d}$ be independent random variables with equation \eqref{equ:approximation}. For $Z = \sum_{j=1}^d\psi_j o_j$, we have $E(Z) = \sum_{j=1}^d\psi_j \lambda_j$ and we define $\sigma=\sum_{j=1}^{d} \psi_{j}^{2} \lambda_{j}, \sigma'=\sum_{j=1}^{d} \psi_{j}^{2} $. Then we have

\begin{gather}
\operatorname{Pr}(Z<E(Z)-\epsilon) \leq e^{-\epsilon^{2} / 2 \sigma} \leq
e^{-\epsilon^{2} / 2 \sigma'}\\
\operatorname{Pr}(Z>E(Z)+\epsilon) \leq e^{-\frac{\epsilon^{2}}{2(\sigma+\hat{\psi} \epsilon / 3)}} \leq
e^{-\frac{\epsilon^{2}}{2(\sigma'+\hat{\psi} \epsilon / 3)}}
\end{gather}

where $\hat{\psi} = max\{\psi_{1}, ...\psi_{j},..., \psi_{d}\}$. Note that $E(Z)$ is exactly the output of the SGC's trainable layer, and $Z$ is the output of SpikingGCN after a linear layer. By applying the upper and lower bounds, the failure probability will be at most $p_f = e^{-\frac{\epsilon^{2}}{2(\sigma'+\hat{\psi} \epsilon / 3)}}$. For question (i) mentioned earlier, we guarantee that our spike representations have approximation to SGC's output with at least $1-p_f$ probability. For question (ii), we also employ the $L_2$ regularization and parameter clip operations during our experiments to control the $\sigma'$ and $\hat{\psi}$ respectively, which can further help us optimize the upper and lower bounds.

\section{Experiments} \label{sec:evaluation}
To evaluate the effectiveness of the proposed SpikingGCN, we conduct extensive experiments that focus on four major objectives: (i) semi-supervised node classification on citation graphs, (ii) performance evaluation under limited training data in active learning, (iii) energy efficiency evaluation on neuromorphic chips, and (iv) extensions to other application domains. Due to the limitation of space, we leave the active learning experiments in  Appendix C.2.

%have designed two different experimental setups. First, to apply our basic model on citation networks, we conducted experiments on four publicly available benchmark datasets: Cora, ACM, citeseer, and Pubmed. We then extend our framework in other application domains, including computer vision and social networks.

\subsection{Semi-Supervised Node Classification}
\paragraph{Datasets.} For node classification, we test our model on four commonly used citation network datasets:
Cora, citeseer, ACM, and Pubmed \cite{wang2019heterogeneous}, where nodes and edges represent the papers and citation links. The statistics of the four datasets are summarized in Table \ref{table:dataset_statistics}. Sparsity refers to the number of edges divided by the square of the number of nodes. 
\begin{table}[h]
\centering
\scriptsize
\begin{tabular}{c|ccccc} \toprule[1pt]
Datasets & Nodes & Edges & Attributes & Classes & Sparsity \\ \midrule[0.5pt]
Cora & $2,708$ & $5,429$ & $1,433$ & $7$ & $0.07\%$ \\
ACM & $3,025$	& $13,128$ & $1,870$ & $3$ &$0.14$\% \\
citeseer & $3,312$ & $4,715$ & $3,703$ & $6$ & $0.04\%$ \\
Pubmed & $19,717$ & $44,324$ & $500$ & $3$ & $0.01\%$ \\ \bottomrule[1pt]
\end{tabular}
\caption{Statistics of the citation network datasets.}
\label{table:dataset_statistics}
\end{table}

\begin{table*}[tb]
\centering
\scriptsize
% \subtable[Table 1]{
\renewcommand{\arraystretch}{1.3}
\begin{tabular}{c|c|c|c|c}
\multicolumn{5}{c}{}\\\toprule[1pt]
 \quad & \textbf{Cora} & \textbf{ACM} & \textbf{citeseer} & \textbf{Pubmed} \\
Models&Split I \qquad  Split II&Split I \qquad  Split II&Split I \qquad  Split II&Split I \qquad  Split II\\\midrule[0.5pt]
% \midrule[0.5pt]
\textbf{GCN}     & $81.0 \pm 1.3 $\quad $87.8 \pm 0.5 $ & $90.0 \pm 0.9$\quad $94.2 \pm 0.6$ & $69.8 \pm 1.6$\quad $73.5 \pm 0.5$& $78.4 \pm 0.5$\quad $87.4 \pm 0.08$ \\
\textbf{SGC}     & $81.5 \pm 0.4$\quad$86.7 \pm 0.8$  & $90.5 \pm 0.9$\quad $93.62 \pm 0.3$ & $71.7 \pm 0.4$\quad$73.06 \pm 0.2$ & $\mathbf{79.2 \pm 0.3}$\quad$86.52 \pm 1.6$  \\
\textbf{FastGCN} & $80.6 \pm 1.2 $\quad$86.5 \pm 0.6 $ & $\mathbf{91.0 \pm 0.7}$\quad$93.85 \pm 0.5$ & $70.0 \pm 0.9$\quad$73.73 \pm 0.9$ & $77.6 \pm 0.6$\quad$88.32 \pm 0.2$ \\
\textbf{GAT}     & $\mathbf{82.5 \pm 0.7}$\quad$87.2 \pm 0.5$  & $90.9 \pm 0.8$\quad$93.54 \pm 0.6$ & $71.6 \pm 0.5$\quad$74.72 \pm 0.7$ & $77.5 \pm 0.5$\quad$86.68 \pm 0.2$ \\
\textbf{DAGNN}   & $\mathbf{84.2 \pm 0.7}$\quad$\mathbf{89.70 \pm 0.1}$  & $\mathbf{91.2 \pm 0.2}$\quad$94.25 \pm 0.3$ & $\mathbf{72.9 \pm 0.2}$\quad$74.66 \pm 0.5$ & $\mathbf{79.8 \pm 0.3}$\quad$87.30 \pm 0.1$  \\
\textbf{SpikingGCN}    & $80.7 \pm 0.6$\quad$88.7 \pm 0.5$  & $ 89.5\pm 0.2$\quad$\mathbf{94.36 \pm 0.2}$ & $72.5 \pm 0.2$\quad$\mathbf{77.56 \pm 0.2}$ & $77.6 \pm 0.5$\quad$\mathbf{89.33 \pm 0.2}$ \\ 
\textbf{SpikingGCN-N}   & $81.0 \pm 0.4$\quad$\mathbf{88.7 \pm 0.1}$  & $90.7 \pm 0.2$\quad$\mathbf{94.78 \pm 0.2}$ & $\mathbf{72.9 \pm 0.1}$\quad$\mathbf{77.80 \pm 0.1}$ & $78.5 \pm 0.2$\quad$\mathbf{89.27 \pm 0.2}$ \\\bottomrule[1pt]
\end{tabular}
\caption{Test accuracy (\%) comparison of different methods. The results from the literature and our experiments are provided. The literature statistics of ACM datasets are taken from \protect\cite{DBLP:conf/kdd/0017ZB0SP20}. All results are  averaged over 10 runs. The top 2 results are boldfaced.}
\label{table:overall_performance}
\end{table*}
\paragraph{Baselines.}\label{sec:baselines}
We implement our proposed SpikingGCN and the following competitive baselines: GCNs \cite{kipf2016semi}, SGC \cite{wu2019simplifying}, FastGCN \cite{chen2018fastgcn}, GAT \cite{velivckovic2017graph}, DAGNN \cite{liu2020towards}. We also conduct the experiments on SpikingGCN-N, a variant of SpikingGCN, which uses a refined Heaviside activation function \eqref{equ:new_heaviside} instead.
%\textbf{Experiment Setup.}
For a fair comparison, we partition the data using two different ways. The first is the same as \cite{DBLP:conf/icml/YangCS16}, which is adopted by many existing baselines in the literature. In this split method (\ie Split I), 20 instances from each class are sampled as the training datasets. In addition, 500 and 1000 instances are sampled as the validation and testing datasets respectively. For the second data split (\ie Split II), the ratio of training to testing is 8:2, and $20\%$ of training samples is further used for validation. %We employ the same dataset split and training procedure for all models on each dataset. 

Table \ref{table:overall_performance} summarizes the node classification's accuracy comparison with the competing methods over four datasets. We show the best results we can achieve for each dataset and have the following key observations: \emph{SpikingGCN achieves or matches SOTA results across four benchmarks on these two different dataset split methods.} It is worth noting that, when the dataset is randomly divided proportionally and SpikingGCN obtains enough data, it can even outperform the state-of-the-art approaches.
For example, SpikingGCN-N outperforms DAGNN by over $3.0\%$ on citeseer dataset. The detailed discussion of the performance can be found in Appendix C.1.

\subsection{Energy Efficiency on Neuromorphic Chips}
To examine the energy efficiency of SpikingGCN, we propose two metrics: i) the number of operations required to predict a node on each model, and ii)  the energy consumed by SpikingGCN on neuromorphic hardware versus other models on GPUs. In this experiment, only the basic SpikingGCN is conducted to evaluate the energy efficiency. The reason we omit SpikingGCN-N is that the negative spikes cannot be implemented on neuromorphic hardware.

We note that training SNN models directly on neuromorphic chip is rarely explored (\cite{thiele2019spikegrad}). In that case, we employ the training phase on GPUs and estimate the energy consumption of test phase on neuromorphic hardware. More importantly, a specific feature of semi-supervised on GNNs is that test data is also visible during the training process. Therefore, the convolutional part during the training covers the global graph. Then during the test phase, no MAC operation is required by our SNN model because all of the data has been processed on GPUs.

%\subsubsection{Operation statistics} 
Estimating the computation overhead relies on operations in the hardware \cite{merolla2014million}. The operation unit of ANNs in contemporary GPUs is usually set to multiply-accumulate (MAC), and for SNNs in the neuromorphic chip is the synaptic operation (SOP). Furthermore, SOP is defined as the change of membrane potential (\ie voltages) in the LIF nodes, and specific statistics in the experiment refer to voltages' changes during charge and fire processes. Following the quantification methods introduced in \cite{hunger2005floating} and ensuring the consistency between different network constraints, we compute the operations of baselines and SpikingGCN to classify one node. Table \ref{table:operation_performance} shows that SpikingGCN has a significant operand reduction. According to the literature  \cite{hu2018spiking,kim2020spiking}, SOPs consume far less energy than MACs, which further highlights the energy efficiency of SpikingGCN.

\begin{table}[h]
\scriptsize
\centering
\begin{tabular}[t]{c|cccc} \toprule[1pt]
models & Cora & ACM & citeseer & Pubmed \\ \midrule[0.5pt]
GCN & $67.77$K & $63.71$K & $79.54$K & $414.16$K \\
SGC & $10.03$K & $5.61$K  & $22.22$K & $1.50$K  \\
FastGCN & $67.54$K & $71.97$K & $141.69$K & $94.88$K \\
GAT & $308.94$K & $349.91$K & $499.16$K & $1.53$M \\
DAGNN & $281.73$K & $210.63$K & $436.11$K & $623.71$K  \\
SpikingGCN & $1.39$K & $0.59$K & $1.19$K  & $0.59$K \\ \bottomrule[1pt]
\end{tabular}
\caption{Operations comparison}
\label{table:operation_performance}
\hfill
\centering
\begin{tabular}[t]{cccc|c} \hline
\toprule[1pt]
\multicolumn{5}{c}{\textbf{GCN on TITAN}}\\
\midrule[0.5pt]
Power (W) & GFLOPS & Nodes & FLOPS & Energy (J) \\\midrule[0.5pt]
280 & 16,310 & 10,000  & 4.14E+09 & 0.07 \\ \midrule[1pt]
\multicolumn{5}{c}{\textbf{SpikingGCN on ROLLs}}\\\midrule[0.5pt]
Voltage (V) & Energy/spike (pJ) & Nodes & Spikes & Energy  \\\midrule[0.5pt]
1.8 & 3.7 & 10,000 & 2.73E+07  & \textbf{1.01E-04} \\ 
\bottomrule[1pt]\hline
\end{tabular}
\caption{Energy consumption comparison}
\label{table:enery consumption}
\end{table}

% \sout{Another quantification indicator about model's efficiency is energy consumption. Note that the SOP is not in direct proportion to the physical energy consumption. That is, based on our empirically analysis, }

However, the energy consumption measured by SOPs may be biased, \eg the zero spikes would also result in the voltage descending changes, which does not require new energy consumption in neuromorphic chips \cite{indiveri2015neuromorphic}. Hence, calculating energy cost only based on operations may result in an incorrect conclusion. To address this issue, we further provide an alternative estimation approach as follow. 
%\subsubsection{Energy cost estimation} 
Neuromorphic designs could provide event-based computation by transmitting one-bit spikes between neurons. This characteristic contributes to the energy efficiency of SNNs because they consume energy only when needed \cite{esser2016convolutional}. 
For example, during the inference phase, the encoded sparse spike trains act as a low-precision synapse event, which costs the computation memory once spikes are sent from a source neuron. 
Considering the above hardware characteristics and the deviation of SOPs in consumption calculation, we follow the spike-based approach utilized in \cite{cao2015spiking} and count the overall spikes during inference for 4 datasets, to estimate the SNN energy consumption.
We list an example of energy consumption when inferring 10,000 nodes in the Pubmed dataset, as shown in Table \ref{table:enery consumption}.

Applying the energy consumed by each spike or operation, in Appendix C.3, we visualize the energy consumption between SpikingGCN and GNNs when employed on the recent neuromorphic chip (ROLLS \cite{indiveri2015neuromorphic}) and GPU (TITAN RTX, 24G \footnote{\url{https://www.nvidia.com/en-us/deep-learning-ai/products/}}), respectively.  Fig. 6 shows that SpikingGCN could use remarkably less energy than GNNs when employed on ROLLs. For example, SpikingGCN could save about 100 times energy than GCN in all datasets. Note that different from GPUs, ROLLS is firstly introduced in 2015, and higher energy efficiency of SpikingGCN can be expected in the future.

\subsection{Extension to Other Application Domains}
In the above experiments, we adopt a basic encoding and decoding process, which can achieve competitive performance on the citation datasets. %This kind of simple but effective prediction on unknown labels arises from the local integration of features. 
However, some other graph structures like image graphs and social networks can not be directly processed using graph Laplacian regularization (\ie \cite{kipf2016semi,wu2019simplifying}). To tackle the compatibility issue, we extend our model and make it adapt to the  graph embedding methods (\ie \cite{DBLP:conf/icml/YangCS16}). 
% Different from the graph Laplacian regularization methods like GCNs, the graph embedding methods always contain specific trainable parameters to incorporate the attributes in the graph structure. In this case, the Bernoulli encoder is unable to generate the spike trains, which perfectly represent the graph information. Taking the image graph as an example, we can see that the Bernoulli encoder cannot fully represent the pixels. Hence, the characteristics of the pixels' local Euclidean neighborhoods must be aggregated. 
We propose a trainable spike encoder, to allow deeper SNNs for different tasks, including classification on grid images and superpixel images, and rating prediction in recommender systems. Limited by space, we leave the implementation detail to Appendix C.4.

\paragraph{Result on Grid Images.} 
To validate the performance of SpikingGCN on image graphs, we first apply our model to the MNIST dataset \cite{lecun1998gradient}. The classification results of grid images on MNIST are summarized in Table \ref{table:mnist_performance}. We choose several SOTA algorithms including ANN and SNN models, which work on MNIST datasets. The depth is calculated according to the layers including trainable parameters. Since we are using a similar network structure as the Spiking CNN \cite{DBLP:journals/corr/LeeDP16}, the better result proves that our clock-driven architecture is able to capture more significant patterns in the data flow. The competitive performance of our model on image classification also proves that SpikingGCN's compatibility to different graph scenarios.

\begin{table}[h]
\scriptsize
\centering
\setlength{\tabcolsep}{4mm}{
\begin{tabular}{c|ccc} \toprule[1pt]
Models &Type& Depth & Accuracy \\ \midrule[0.5pt]
SplineCNN \cite{DBLP:conf/cvpr/FeyLWM18} &ANN    & $8$ & $99.22$\\
 LeNet5 \cite{lecun1998gradient}   &ANN  & $4$  & $99.33$\\
 LISNN \cite{DBLP:conf/ijcai/ChengHX020}  &SNN   & $6$  & $99.50$\\
Spiking CNN \cite{DBLP:journals/corr/LeeDP16} &SNN & $4$  & $99.31$ \\
S-ResNet \cite{hu2018spiking} &SNN  & $8$  & $\mathbf{99.59}$ \\ 
SpikingGCN (Ours)   &SNN  & $4$  & $99.35$\\\bottomrule[1pt]
\end{tabular}
\caption{Test accuracy (\%) comparison on MNIST. The best results are boldfaced.}
\label{table:mnist_performance}}
\end{table}

\paragraph{Results on Superpixel Images.} We select the MNIST superpixel dataset \cite{DBLP:conf/cvpr/MontiBMRSB17} for the comparison with the grid experiment mentioned above. The results of the MNIST superpixel experiments are presented in Table \ref{table:superpixel_performance}. 
Since our goal is to prove the generalization of our model on different scenarios, we only use 20 time steps to conduct this subgraph classification task and achieve the mean accuracy of $94.50\%$ over 10 runs. It can be seen that SpikingGCN is readily compatible with the different convolutional methods of the graph and obtain a competitive performance through a biological mechanism.  
\begin{table}[h]
\centering
\scriptsize
\setlength{\tabcolsep}{10mm}{
\begin{tabular}{c|c} \toprule[1pt]
Models & Accuracy\\\midrule[0.5pt]
ChebNet \cite{DBLP:conf/nips/DefferrardBV16}&$75.62$\\
MoNet \cite{DBLP:conf/cvpr/MontiBMRSB17}&$91.11$\\
SplineCNN \cite{DBLP:conf/cvpr/FeyLWM18}&$\mathbf{95.22}$\\
SpikingGCN (Ours) &$94.50$\\ \bottomrule[1pt]
\end{tabular}
\caption{Test accuracy comparison on MNIST. The best results are boldfaced. Baseline numbers are taken from \protect\cite{DBLP:conf/cvpr/FeyLWM18}.}
\label{table:superpixel_performance}
}
\end{table}
\begin{table}[h]
\scriptsize
\centering
\setlength{\tabcolsep}{8mm}{
\begin{tabular}{c|c} \toprule[1pt]
Models & RMSE Score\\\midrule[0.5pt]
MC \cite{DBLP:journals/cacm/CandesR12}&$0.973$\\
GMC \cite{DBLP:journals/corr/KalofoliasBBV14}&$0.996$\\
GRALS \cite{DBLP:conf/nips/RaoYRD15}&$0.945$\\
sRGCNN \cite{DBLP:conf/nips/MontiBB17}&$0.929$\\
GC-MC \cite{DBLP:journals/corr/BergKW17}&$\mathbf{0.910}$\\ 
SpikingGCN (Ours)&$0.924$\\\bottomrule[1pt]
\end{tabular}
\caption{Test RMSE scores with MovieLens 100K datasets. Baselines numbers are taken from \protect\cite{DBLP:journals/corr/BergKW17}.}
\label{table:recom_performance}
}
\end{table}

\paragraph{Results on Recommender Systems.}
 We also evaluate our model with a rating matrix extracted from MovieLens 100K \footnote{\url{https://grouplens.org/datasets/movielens/}} and report the RMSE scores compared with other matrix completion baselines in Table \ref{table:recom_performance}. The comparable loss $0.924$ indicates that our proposed framework can also be employed in recommender systems. Because the purpose of this experiment is to demonstrate the applicability of SpikingGCN in recommender systems, we have not gone into depth on the design of a specific spike encoder. We leave this design in the future work since it is not the focus of the current paper.

\section{Conclusions}
In this paper, we present SpikingGCN, a first-ever bio-fidelity and energy-efficient framework focusing on graph-structured data, which encodes the node representation and makes the prediction with less energy consumption. In our basic model for citation networks, 
% the encoded spike trains are processed by a simple linear layer combined with a neuron layer. W
we conduct extensive experiments on node classification with four public datasets. Compared with other SOTA approaches, we demonstrate that SpikingGCN achieves the best accuracy with the lowest computation cost and much-reduced energy consumption. Furthermore, SpikingGCN also exhibits great generalization when confronted with limited data. In our extended model for more graph scenarios, SpikingGCN also has the potential to compete with the SOTA models on tasks from computer vision or recommender systems. Relevant results and discussions are presented to offer key insights on the working principle, which may stimulate future research on environmentally friendly and biological algorithms.

\section*{Acknowledgments}
The research is supported by the Key-Area Research and Development Program of Guangdong Province (2020B010165003), the Guangdong Basic and Applied Basic Research Foundation (No. 2020A1515010831), the Guangzhou Basic and Applied Basic Research Foundation (No. 202102020881), the Tencent AI Lab RBFR2022017, and the Program for Guangdong Introducing Innovative and Entrepreneurial Teams (No. 2017ZT07X355). Qi Yu is supported in part by an NSF IIS award IIS-1814450 and an ONR award N00014-18-1-2875. The views and conclusions contained in this paper are those of the authors and should not be interpreted as representing
any funding agency.
%\clearpage
% \input{appendix}

\footnotesize

\bibliographystyle{named}

\bibliography{ijcai22}

% \iffalse

\appendix

\begin{center}
    {\bf \large Appendix}
\end{center}
\paragraph{Organization of the Appendix.}  In this Appendix, we first discuss some additional related work that provides a more complete context of the proposed SpikingGCN model. We then describe the detailed training process of the model, which is accompanied by the link to access the entire source code. Finally, we show additional experimental results that complement the ones presented in the main paper. 

\section{Related Work} \label{related work}

\paragraph{Spiking Neural Networks. }
The fundamental SNN architecture includes the encoder, spiking neurons, and interconnecting synapses with trainable parameters \cite{tavanaei2019deep}. These procedures contribute to the substantial integrate-and-fire (IF) process in SNNs: any coming spikes lead to the change of the membrane potential in the neuron nodes; once membrane potentials reach the threshold voltage, the neuron nodes fire spikes and transmit the messages into their next nodes.

Some studies have developed the methodology along with a function to approximate the non-differentiable IF process \cite{jin2018hybrid,zhang2020spike}. Although gradient descent and error back-propagation are directly applicable for SNNs in that way, a learning phase strongly related to ANNs still causes a heavy burden on the computation. Another approach to alleviate the difficulty of training in SNNs is using an ANN-to-SNN conversion by using the pre-trained neuron weights. \cite{kim2020spiking} take advantage of the weights of pre-trained ANNs to construct a spiking architecture for object recognition or detection. Although those conversions can be successfully performed, multiple operators of already trained ANNs are not fully compatible with SNNs \cite{rueckauer2017conversion}. As a result, SNNs constructed from a fully automatic conversion of arbitrary pre-trained ANNs are not able to achieve a comparable prediction performance.

Another popular way to build the SNNs models is the spike-timing-dependent-plasticity (STDP) learning rule, where the synaptic weight is adjusted according to the interval between the pre- and postsynaptic spikes. \cite{DBLP:journals/ficn/Diehl015} propose an unsupervise learning model, which utilizes more biologically plausible components like conductance-based synapses and different STDP rules to achieve competitive performance on the MNIST dataset. \cite{lee2018training} introduce a pre-training scheme using biologically plausible unsupervised learning to better initialize the parameters in multi-layer systems. Although STDP models provide a closer match to biology for the learning process, how to achieve a higher level function like classification using supervised learning is still unsolved \cite{cao2015spiking}. Besides, it can easily suffer from prediction performance degradation compared with supervised learning models.

\paragraph{Graph neural networks.} Unlike a standard neural network, GNNs need to form a state that can extract the representation of a node from its neighborhood with an arbitrary graph \cite{liu2020abstract}. In particular, GNNs utilize extracted node attributes and labels in graph networks to train model parameters in a specific scenario, such as citation networks, social networks, protein-protein interactions (PPIs), and so on. GAT \cite{velivckovic2017graph} has shown that capturing the weight via an end-to-end neural network can make more important nodes receive larger weights.
In order to increasingly improve the accuracy and reduce the complexity of GCNs, the extended derivative SGC \cite{wu2019simplifying} eliminates the nonlinearities and collapses weight matrices between consecutive layers. FastGCN \cite{chen2018fastgcn} successfully reduces the variance and improves the performance by sampling a designated number of nodes for each convolutional layer. Nonetheless, these convolutional GNN algorithms rely on high-performance computing systems to achieve fast inference for high-dimensional graph data due to a heavy computational cost. Since GCNs bridge the gap between spectral-based and spatial-based approaches \cite{xie2020graph}, they offer desirable flexibility, extensibility, and architecture complexity. Thus, we adopt the GCN-based feature processing to construct our basic SNNs model.

\paragraph{Energy consumption estimation.}
An intuitive measurement of the model's energy consumption is investigating the practical electrical consumption. \cite{DBLP:conf/acl/StrubellGM19} propose to repeatedly query the NVIDIA System Management Interface \footnote{nvidia-smi: \url{https://bit.ly/30sGEbi}} to obtain the average energy consumption for training deep neural networks for natural language processing (NLP) tasks. \cite{DBLP:journals/corr/CanzianiPC16} measure the average power draw required during inference on GPUs by using the Keysight 1146B Hall effect current probe. However, querying the practical energy consumption requires very strict environment control (e.g., platform version and temperature), and might include the consumption of background program, which results in the inaccuracy measurement. Another promising approach to estimate the model's energy consumption is according to the operations during training or inference.  \cite{DBLP:journals/corr/abs-2007-03051} develop a tool for calculating the operations of different neural network layers, which helps to track and
predict the energy and carbon footprint of ANN models. Some SNN approaches \cite{hu2018spiking,kim2020spiking} successfully access the energy consumed by ANN and SNN models by measuring corresponding operations multiplied by theoretical unit power consumption. This kind of methods can estimate the ideal energy consumption excluding environmental disturbance.
In addition, contemporary GPU platforms are much more mature than SNN platforms or neuromorphic chips \cite{merolla2014million,indiveri2015neuromorphic}. As a result, due to the technical restriction of employing SpikingGCN on neuromorphic chips, we theoretically estimate the energy consumption in the experimental section.

\section{Notation, algorithm and source code} \label{algorithm}
We list the frequently used notation in Table \ref{table:fre_notations}.
Algorithm~\ref{alg:example} shows the detailed training process of the proposed SpikingGCN model. The source code can be accessed via \url{https://anonymous.4open.science/r/SpikingGCN-1527}.
\begin{table}[h]
\centering
\caption{Frequently used notations in this paper}
\label{table:fre_notations}
\renewcommand{\arraystretch}{1.2}
\scriptsize
\setlength{\tabcolsep}{3mm}{
\begin{tabular}{c|c} \toprule[1pt]
Notations & Descriptions\\\hline \hline
$\mathcal{G}$&Graph structure data\\\hline
$v_i$&Single node in the graph\\\hline
$\mathcal{V}$&Node set in the graph\\\hline
$N, C, d$& Number of nodes, class number and feature dimensions\\\hline
$a_{ij}$& Edge weight between nodes $v_i$ and $v_j$, scalar\\\hline
$\bm{A}$& Adjacent matrix of the graph, $\mathbb{R}^{N\times{N}}$\\\hline
$x_i$&Feature vector of $i$-th node, $\mathbb{R}^{1\times{d}}$ \\\hline
$\bm{X}$& Entire attribute matrix in the graph, $\mathbb{R}^{N\times{d}}$ \\\hline
$Y$& One-hot labels for each node, $\mathbb{R}^{N\times{C}}$\\\hline
$d_i$&Degree of a single node,scalar\\\hline
$\bm{D}$&Diagonal matrix of the degree of each node, $\mathbb{R}^{N\times{N}}$\\\hline
$h_i$&New feature of $i$-th node after convolution, $\mathbb{R}^{1\times{d}}$\\\hline
$\bm{H}$&Entire attribute matrix after convolution, $\mathbb{R}^{N\times{d}}$\\\hline
$T$&Time step in the clock-driven SNNs\\\hline
$O^{pre}$&Spike of one node generated by encoder , $\mathbb{R}^{1\times{d}}$\\\hline
$O^{post}$&Spike of one node generated by decoder, $\mathbb{R}^{1\times{C}}$\\\hline
$\lambda_i$&$j$-th feature value of a single node, scalar\\\hline
$o_j$&Basic spike unit generated by $j$-th feature value, equal to 0 or 1\\\hline
$V_m^t$&Membrane potential at $t$-th time step in the decoder, $\mathbb{R}^{1\times{C}}$\\\hline
$\mathbf{\psi}$&Trainable weight matrix, $\mathbb{R}^{d\times{C}}$\\\hline
$\tau_m$&Time  constant, hyperparameter, scalar\\\hline
$V_{reset}$&Signed reset voltage, hyperparameter, scalar\\\hline
$V_{th}$&Spiking threshold, hyperparameter, scalar\\\bottomrule[1pt]
\end{tabular}}
\end{table}

\begin{algorithm}[htb]
   \caption{Model Training of SpikingGCN}
   \label{alg:example}
   \textbf{Input:} Graph $\mathcal{G}(\mathcal{V},\mathbf{A})$; input attributes $x_i\in\mathbf{X}$; one-hot matrix of label $y_i\in \mathbf{Y}_o$; \\
   \textbf{Parameter:} Learning rate $\beta$; Weight matrix $\mathbf{w}$; embedding function \textsc{embedding()}; encoding function \textsc{encoding()}; \\
   charge, fire, reset functions \textsc{charge()}, \textsc{fire()}, \textsc{reset()} \\
   \textbf{Output:} Firing rate vector $\hat{y}_i$ for training subset $\mathcal{V}_o$, which is the prediction
   
\begin{algorithmic}[1]  
   \WHILE{not converge}
   \STATE Sample a mini-batch nodes $\mathcal{V}_{l}$ from the training nodes $\mathcal{V}_o$
   \FOR{each node $i\in\mathcal{V}_{l}$}
   \STATE $h_i\gets\textsc{embedding}(\mathbf{A}, x_i)$ // Eq. (\ref{equ:conv1})(\ref{equ:conv2})
   \FOR{$t=1...T$}
   \STATE $O^{pre}_{i,t}\gets\textsc{encoding}(h_i)$ // Eq. (\ref{encoding})
   \STATE $V_m^t = \textsc{charge}(\mathbf{W}\cdot O^{pre}_{i,t}$) // Eq. (\ref{equ:charge_1})
   \STATE $O^{post}_{i,t} = \textsc{fire}(V_m^t)$
   \STATE $V_m^t = \textsc{reset}(V_m^t)$ // Eq. (\ref{equ:reset})
%   \STATE $O^{t(pos)}_v\gets\textsc{charge}()$ // Eq. (\ref{equ:charge_1}) - Eq. (\ref{equ:charge_2}) 
%   \STATE $y^t_v\gets\textsc{fire}(O^{t(pos)}_v)$
   \ENDFOR
   \STATE $\hat{y}_i\gets\frac{1}{T}\sum{O^{post}_{i,t}}$
   \ENDFOR
   \STATE Perform meta update, $\mathbf{w}\gets\mathbf{w}-\beta\nabla_{\mathbf{w}}\mathcal{L}(y_i,\hat{y}_i)$
   \ENDWHILE
\end{algorithmic}
\end{algorithm}

\section{Additional Experimental Results}
We report additional experimental results that complement the ones reported in the main paper. 

\subsection{Discussion of Node Classification Experiments }\label{citation discussion}

\begin{figure}
  \begin{center}
    \includegraphics[width=7.5cm]{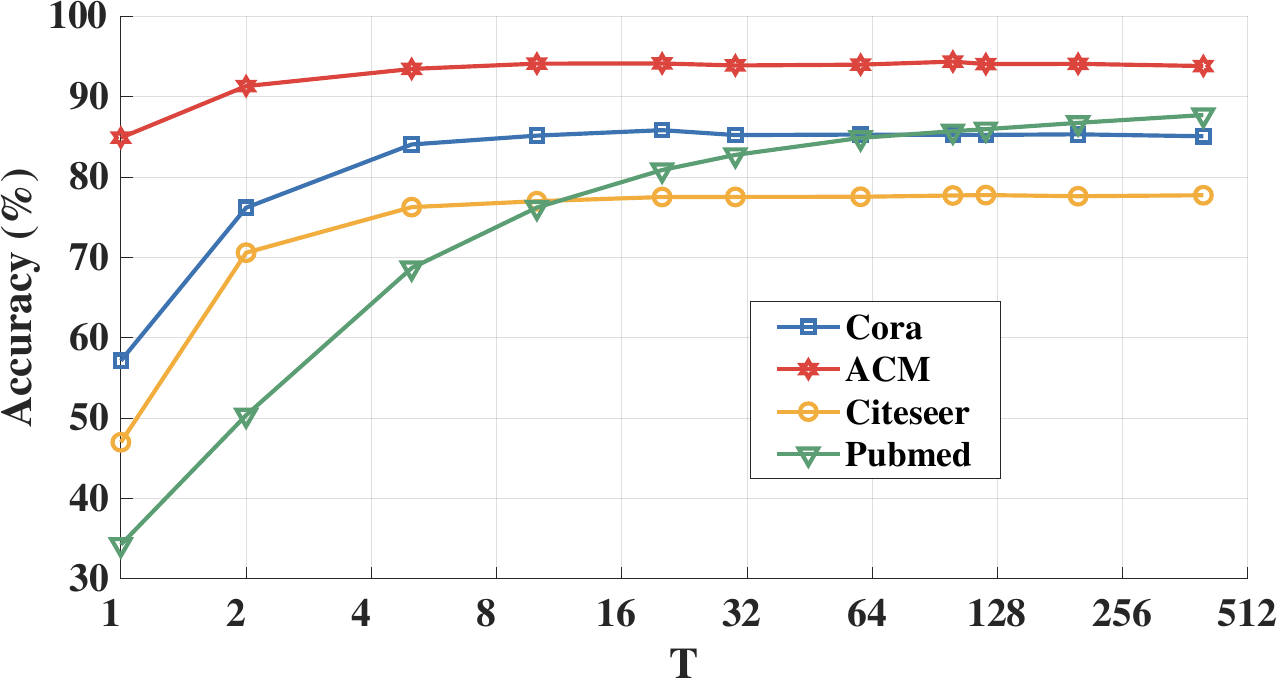}
  \end{center}
  \caption{\label{fig:t}Impact of T}
\end{figure}

The remarkable performance of bio-fidelity SpikingGCN is attributed to three main reasons. First, as shown in Fig. \ref{fig:t}, an appropriate $T$ can enable our network to focus on the most relevant parts of the input representation to make a decision, similar to the attention mechanism \cite{velivckovic2017graph}. Note that an optimal $T$ relies on different statistical patterns in the dataset. In another word, we can also view the Bernoulli encoder as a moderate max-pooling process on the graph features, where the salient representation of each node can have a higher probability to be the input of the network. As a result, assigning varying importance to nodes enable SpikingGCN to perform more effective prediction on the overall graph structure.

\begin{figure}
  \begin{center}
    \includegraphics[width=3.3in]{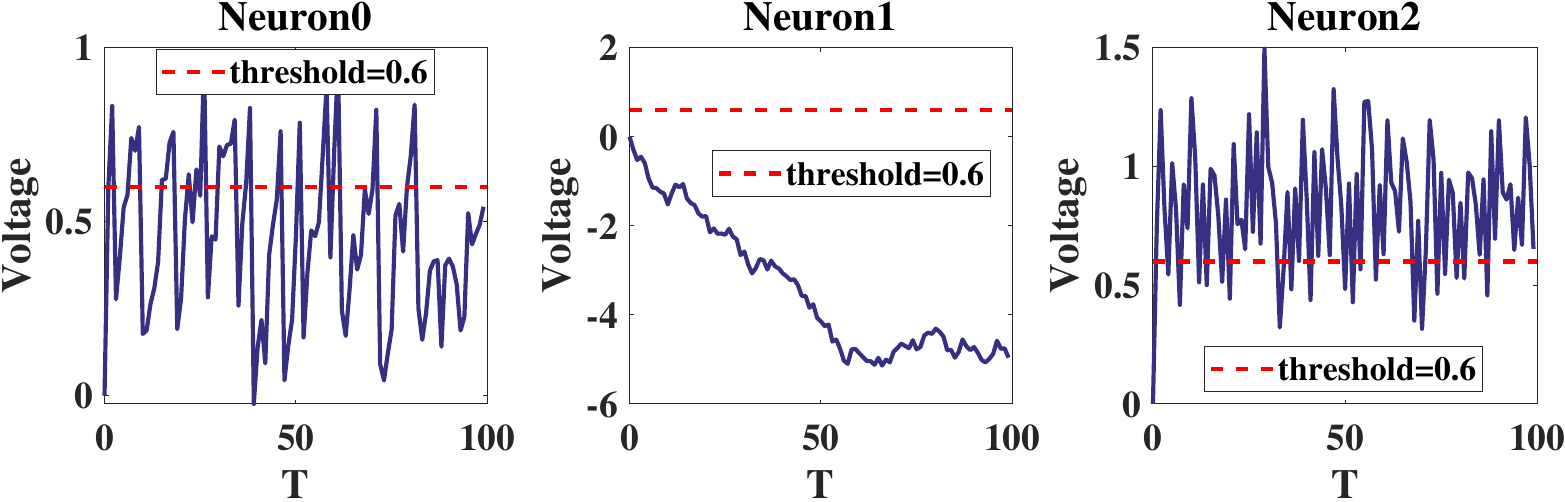}
  \end{center}
  \caption{\label{fig:potential}Membrane potential activity}
\end{figure}
Second, based on our assumption, the majority of accurate predictions benefit from attribute integration. We simplify the network and make predictions using fewer parameters, which effectively reduces the chance of overfitting. The significant performance gain indicates the better generalization ability of neural inference trained with the simplified network, which validates the effectiveness of bio-fidelity SpikingGCN. Last, the variant SpikingGCN-N has achieved better results than the original one on Cora, ACM, and citeseer datasets. As shown in Fig. \ref{fig:potential}, part of the negative voltages will be converted into negative spikes by the Heaviside activation function. The negative spikes can play a role in suppression since the spikes of $T$ times are summed to calculate the fire ratio, which is more biologically plausible. However, the improvement seems to have no effect on Pubmed, which has the highest sparsity and the lowest number of attributes. Sparse input leads to sparse spikes and voltages, and negative spikes tend to provide overly dilute information because the hyperparameters (\eg $-1/\theta$ of Heaviside activation function) are more elusive.

\subsection{SpikingGCN for Active Learning}\label{active learning}

\begin{table}[h]
\centering
\caption{The Area under the Learning Curve (ALC) on Cora and ACM datasets.}
\label{table:active_table}
\scriptsize
\setlength{\tabcolsep}{4.0mm}{
\begin{tabular}{c|cc} \toprule[1pt]
 & Cora & ACM  \\ \midrule[0.5pt]
SOPT-SpikingGCN & $\mathbf{72.9 \pm 0.3}$ & $\mathbf{87.7\pm 0.4}$  \\
SOPT-GCN & $71.3 \pm 0.2$ & $85.8\pm 0.7$  \\\midrule[0.5pt]
PE-SpikingGCN & $\mathbf{62.6\pm 1.1}$ & $\mathbf{85.1\pm 1.3}$  \\
PE-GCN & $59.3 \pm 1.4$ & $83.2 \pm 1.0$  \\\midrule[0.5pt]
Random-SpikingGCN & $\mathbf{60.8\pm 2.0}$ & $\mathbf{84.7\pm 1.7}$  \\
Random-GCN & $57.3\pm 2.1$ & $82.7 \pm 1.5$  \\\bottomrule[1pt]
\end{tabular}}
\end{table}

\begin{figure*}[htbp]
\centering

\subfigure[Active learning on the Cora dataset]{
\begin{minipage}[t]{0.33\linewidth}
\centering
\includegraphics[width=1.8in]{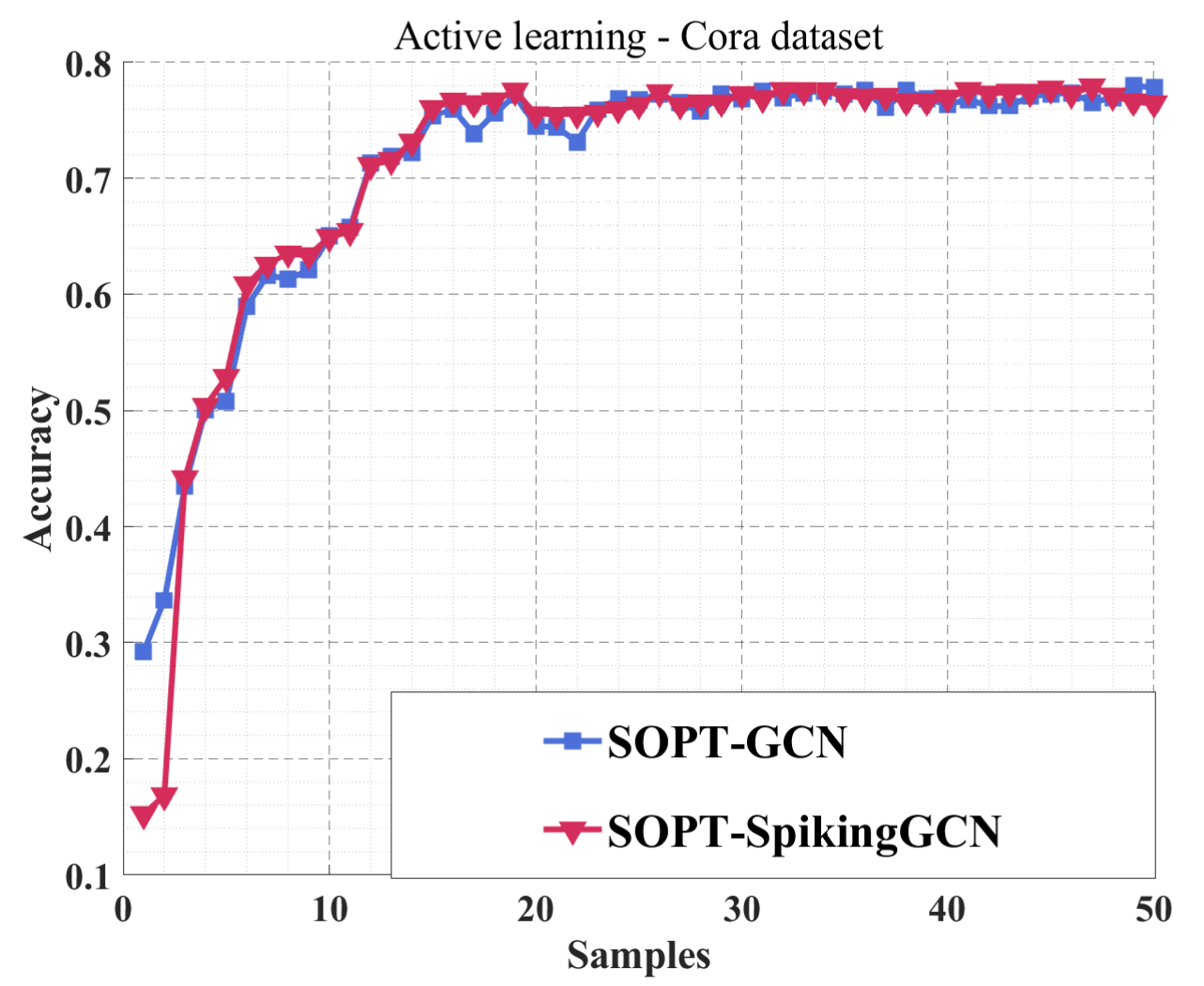}
%\caption{fig1}
\end{minipage}%

\begin{minipage}[t]{0.33\linewidth}
\centering
\includegraphics[width=1.8in]{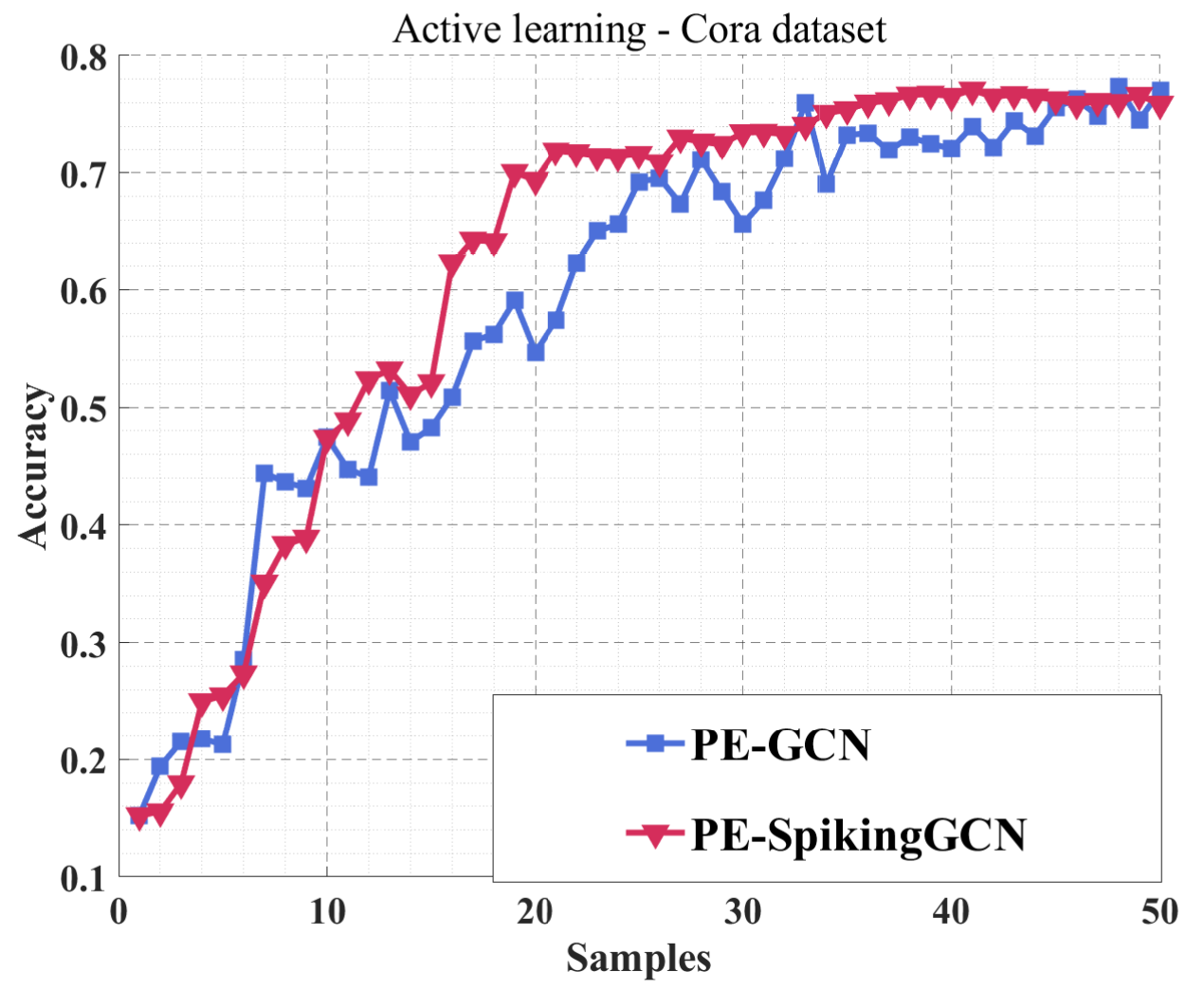}
%\caption{fig2}
\end{minipage}%
\begin{minipage}[t]{0.33\linewidth}
\centering
\includegraphics[width=1.8in]{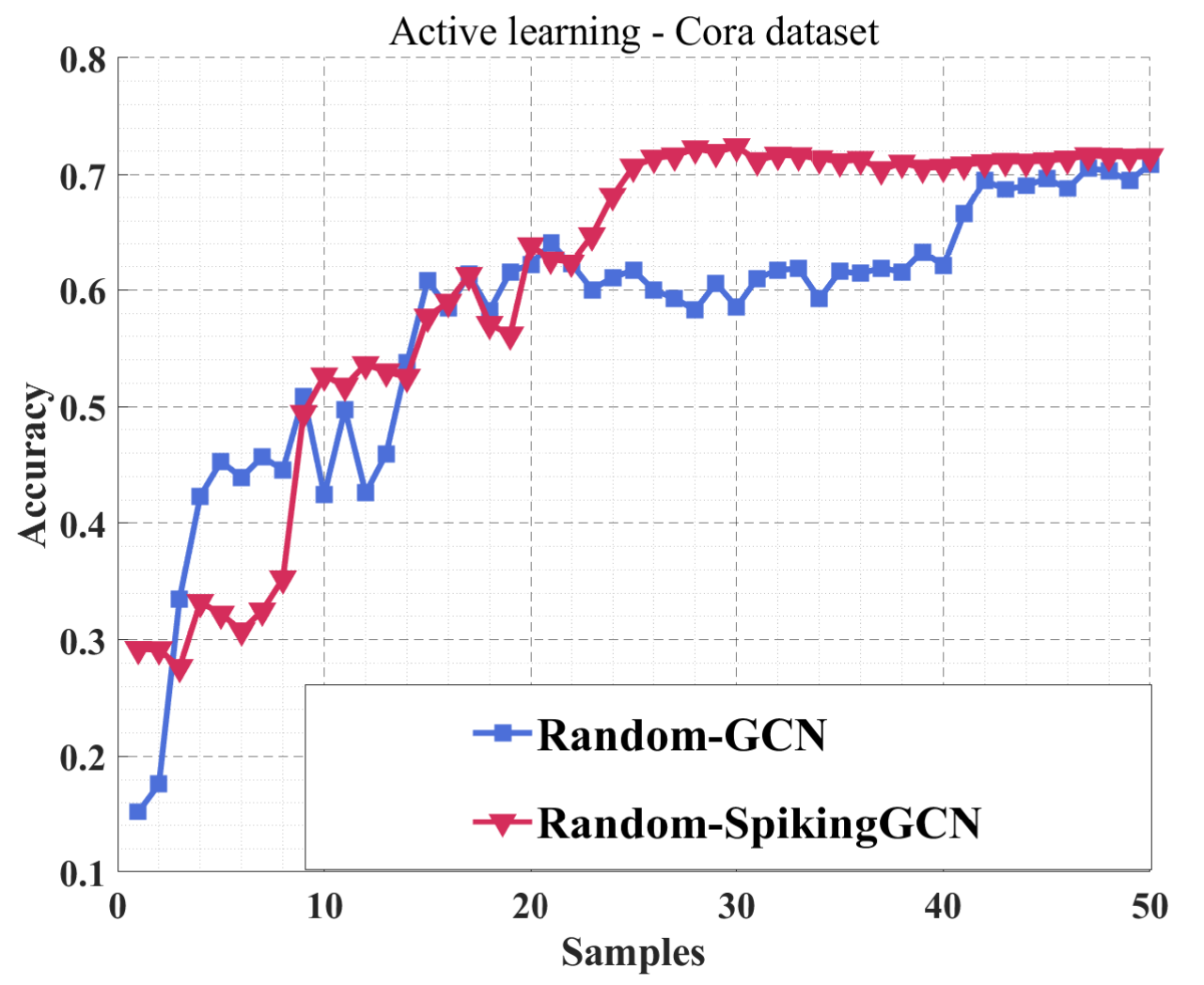}
%\caption{fig1}
\end{minipage}%
}
\centering
\subfigure[Active learning on the ACM dataset]{
\begin{minipage}[t]{0.33\linewidth}
\includegraphics[width=1.8in]{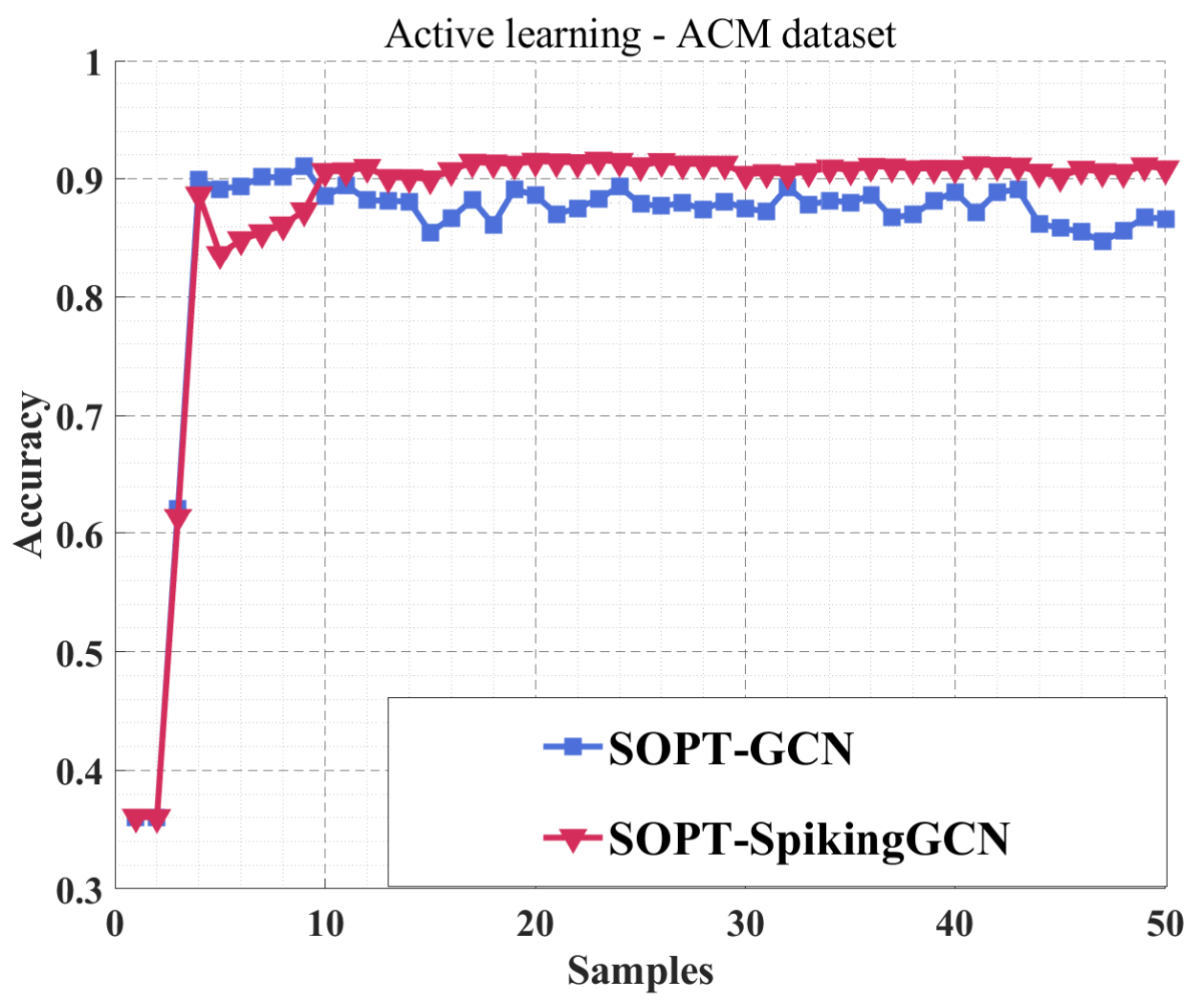}
%\caption{fig2}
\end{minipage}
\begin{minipage}[t]{0.33\linewidth}
\centering
\includegraphics[width=1.8in]{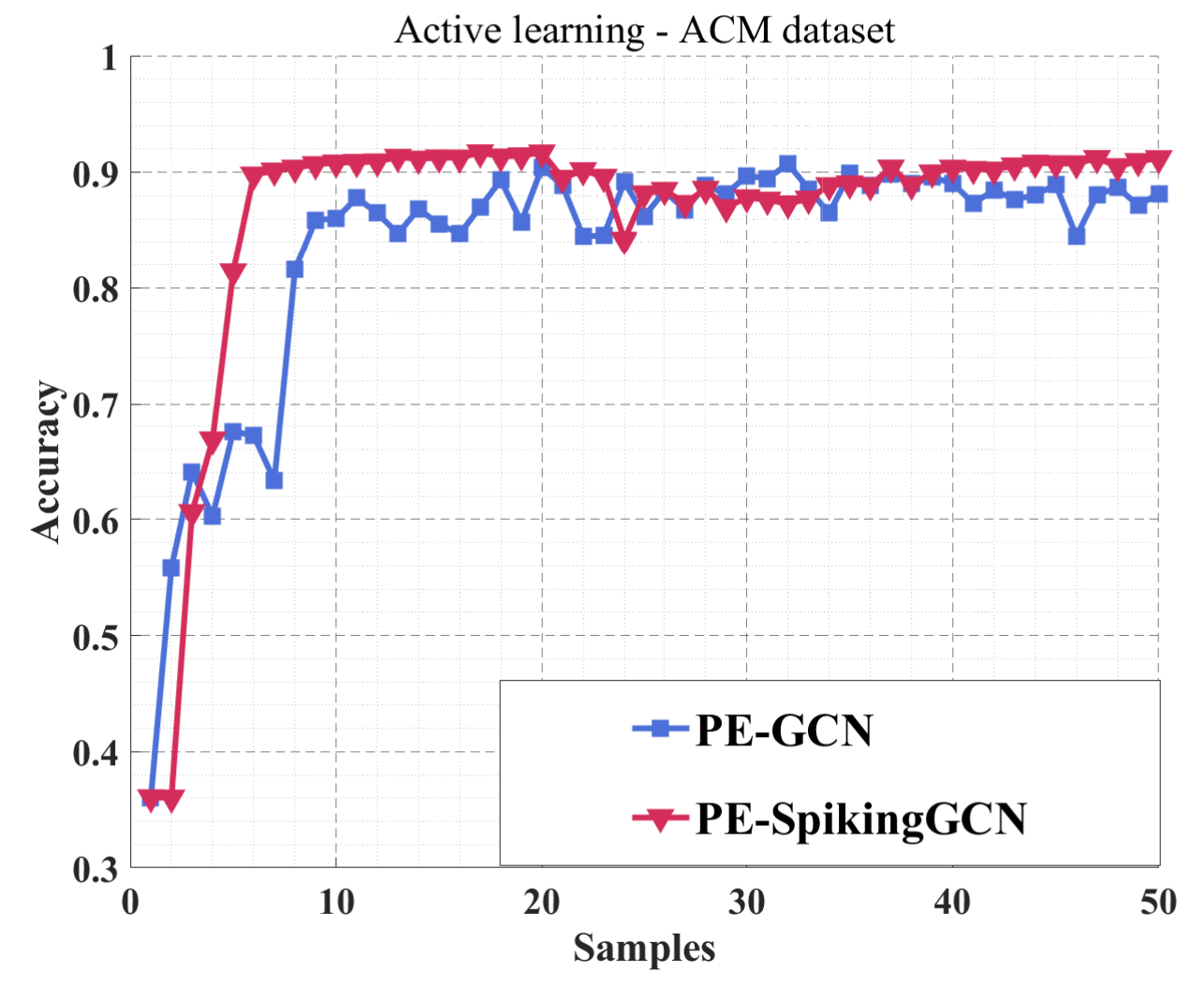}
%\caption{fig2}
\end{minipage}
\begin{minipage}[t]{0.33\linewidth}
\centering
\includegraphics[width=1.8in]{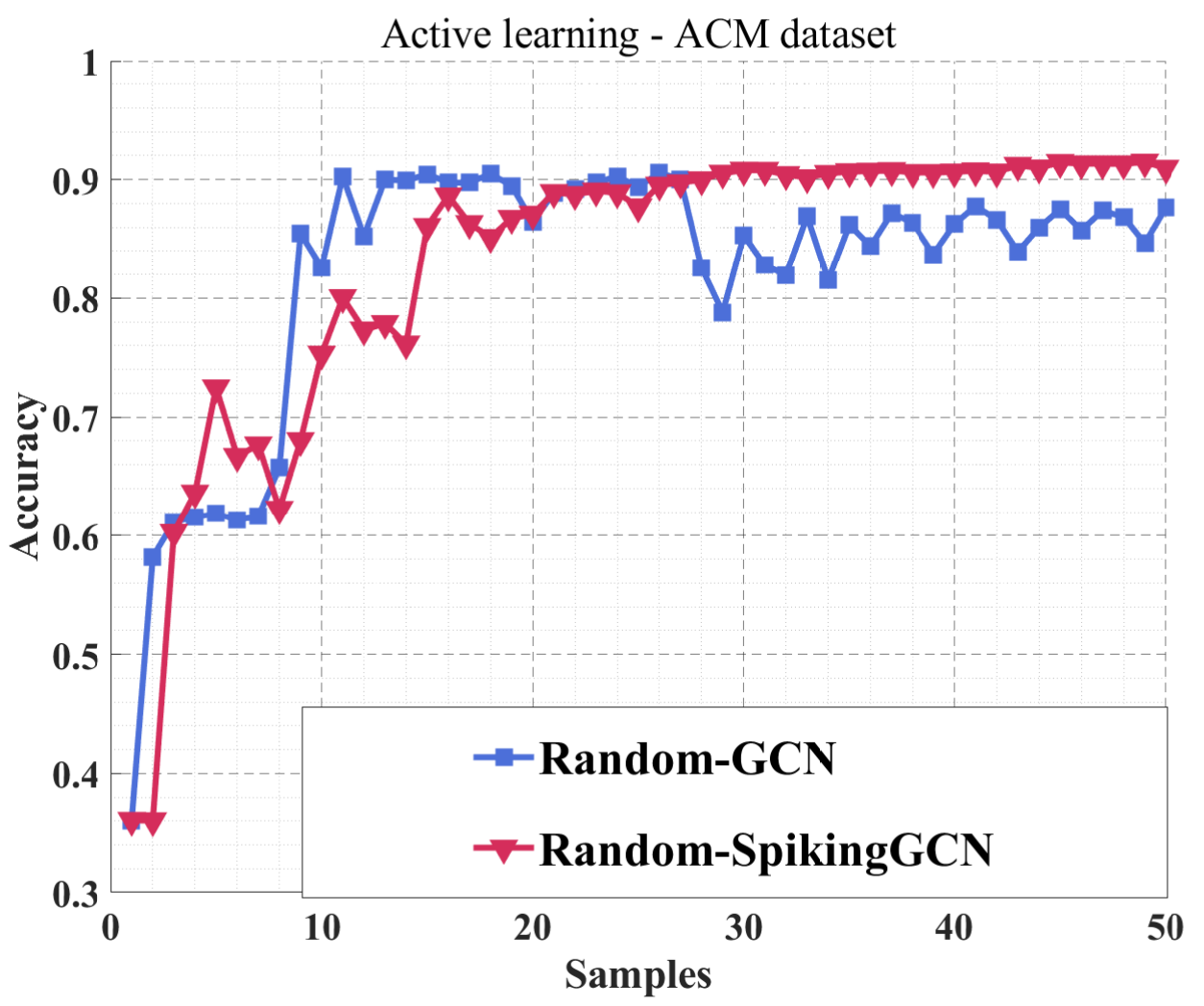}
%\caption{fig1}
\end{minipage}%
}%
\centering
\caption{Active learning curves for both Cora and ACM datasets.}
\label{fig:active_figure}
\end{figure*}
Based on the prediction result above, we are interested in SpikingGCN's performance when the training samples vary, especially when the data is limited.
Active learning has the same problem as semi-supervised learning in that labels are rare and costly to get. The objective of active learning is to discover an acquisition function that can successively pick unlabeled data in order to optimize the prediction performance of the model. Thus, instead of obtaining unlabeled data at random, active learning may help substantially increase data efficiency and reduce cost. Meanwhile, active learning also provides a way to evaluate the generalization capability of models when the data is scarce.
Since SpikingGCN can achieve a $3.0$ percent performance improvement with sufficient data, we are interested in how the prediction performance changes as the number of training samples increases. 

\paragraph{Experiment Setup.}
We apply SpikingGCN and GCN as the active learners and observe their performance. Furthermore, three kinds of acquisition methods are considered. First, according to \cite{DBLP:conf/nips/MaGS13}, the $\sum$- optimal (SOPT) acquisition function is model agnostic because it only depends on the graph Laplacian to determine the order of unlabeled nodes. The second one is the standard predictive entropy (PE) \cite{DBLP:conf/nips/Hernandez-LobatoHG14}. Last, we consider random sampling as the baseline. Starting with only one initial sample, the accuracy is periodically reported until 50 nodes are selected. Results are reported on both Cora and ACM datasets.

%\subsection{Energy Efficiency on Neuromorphic Chips}\label{energy efficiency}

\begin{figure}[htbp]
\centering\includegraphics[width=3.3in]{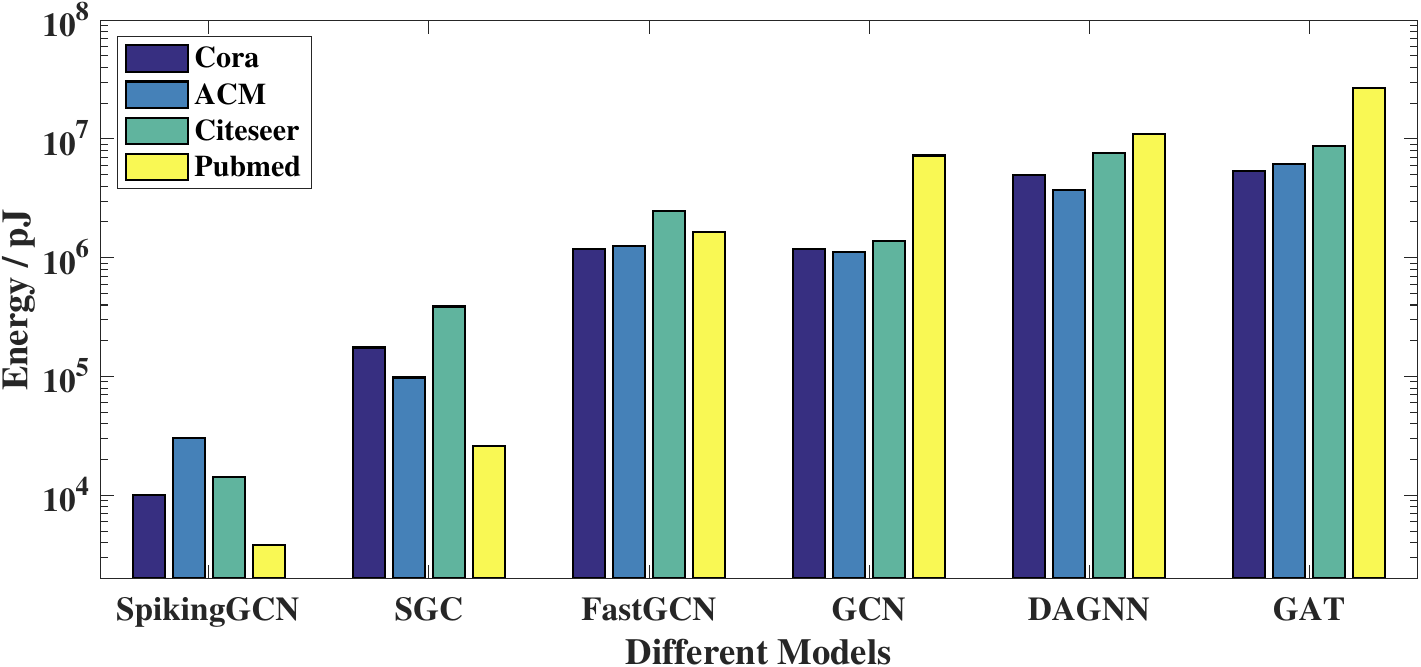}
\caption{The energy consumption of SpikingGCN and baselines on their respective hardware.}\label{fig:energy}
\end{figure}

The Area under the Learning Curve
(ALC) \footnote{ALC corresponds to the area under the learning curve and is constrained to have the maximum value 1.} results are shown in Table \ref{table:active_table}. 
We provide the active learning curves of SpikingGCN and GCN in Fig. \ref{fig:active_figure}, which are consistent with the statistics reported in Table \ref{table:active_table}. 
It can be seen that SOPT can choose the most informative nodes for SpikingGCN and GCN. At the same time, the PE acquisition function is a moderate strategy for performance improvement. Finally, in random strategy both models suffer from high variations during prediction as well as unstable conditions throughout the active learning process. However, no matter which strategy is adopted, SpikingGCN achieves a better generalization than GCN when the training data is scarce.

\subsection{Energy Efficiency Experiments}\label{Energy Efficiency}

Fig. \ref{fig:energy} shows the remarkable energy difference between SpikingGCN and GNN based models. %As for such a significant energy gap between our method and GNNs, we explore other deep SNN models (S-ResNet, Spiking-Yolo, LISNN \cite{hu2018spiking,kim2020spiking,chenglisnn}) and offer two interpretations. i) 
First, the sparse characteristic of graph datasets fits the spike-based encoding method. Furthermore, the zero values in node representations would have no chance to inspire a synapse event (spike) on a neuromorphic chip, leading to no energy consumption. Second, our simplified network architecture only contains two main neuron layers: a single fully connected layer and an LIF layer. Consider Pubmed as an example. Few attributes and a sparse adjacency matrix result in sparse spikes, and the smaller number (\ie 3) of classes also require fewer neurons. This promising results imply that SpikingGCN could have the potential to achieve more significant advantages in energy consumption than general GNNs.

\subsection{SpikingGCN on Other Application Domains}\label{grid images}
\begin{figure}[ht]
\centering
\includegraphics[width=3.2in]{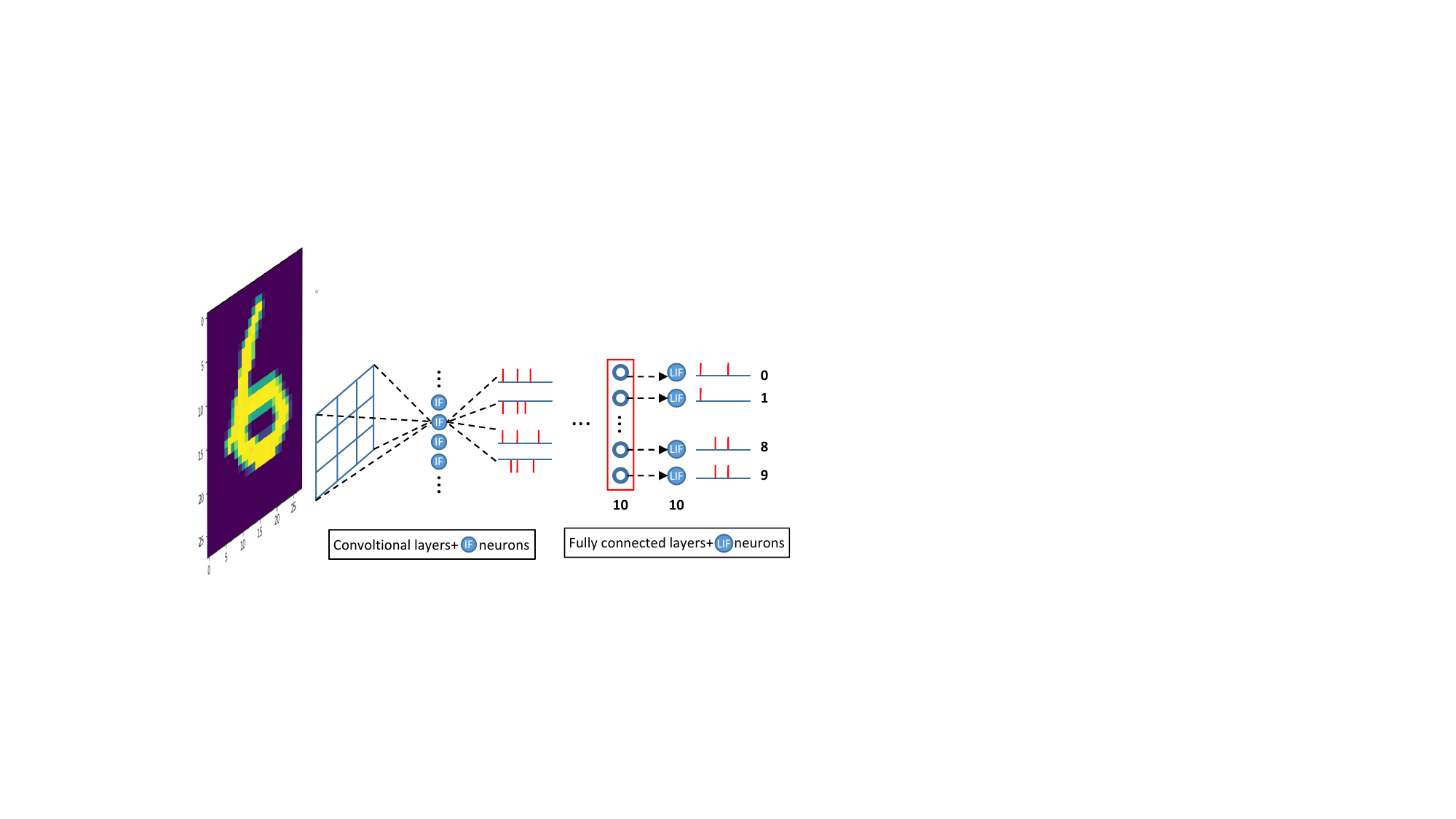} 
\caption{An extended model for deep SNNs}
\label{fig:extended}
\end{figure}

{\bf Results on image grids.} The MNIST dataset contains 60,000 training samples and 10,000 testing samples of handwritten digits from 10 classes. Each image has $28\times28=784$ grids or pixels, hence we treat each image as a node which has 784 features. It is worth noting that the grid image classification is identical to the citation networks where node classes will be identified, with the exception of the absence of an adjacent matrix. %As a common representation in the computer vision area, the experiment on grid images is an ideal choice to validate the generalization of our model on graph-based data. 
To extend our model, we adopt the traditional convolutional layers and provide the trainable spike encoder for graph embedding models, and the extended framework is given by Fig. \ref{fig:extended}. Since the LIF neuron models contain the leaky parameters $\tau_m$, which can decay the membrane potential $V_m^t$ and activate the spikes on a small scale, we adopt the Integrate-and-Fire (IF) process to maintain a suitable firing rate for the encoder. The membrane activity happening in the spike encoder can be formalized as:
\begin{equation}
 V_m^{t}=V_m^{t-1}(1-|\text{H}\left(V_m^{t-1}\right)|)+X(t),
\end{equation}
where $X(t)$ is the convolutional output at time step $t$, and $\text{H}\left(V_m^{t-1}\right)$ is given in \eqref{equ:new_heaviside}. As shown in Fig. \ref{fig:extended}, the convolutional layers combined with the IF neurons will perform an auto-encoder function for the input graph data. After processing the spike trains, the fully connected layers combined with the LIF neurons can generate the spike rates for each class, and we will obtain the prediction result from the most active neurons.

\begin{figure}[h]
\centering
\includegraphics[width=3.2in]{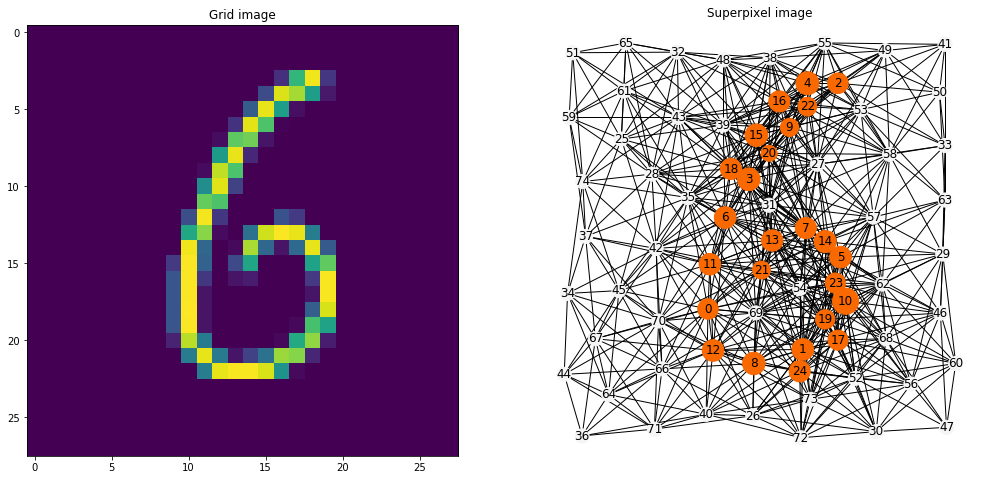} 
\caption{Comparison between grid images and superpixel images}
\label{fig:grid_super}
\end{figure}

\paragraph{Results on superpixel images.}\label{superpixel images}

Another more complex graph structure is the superpixel images. Compared with the general grid images, superpixel images represent each picture as a graph which consists of connected nodes. Hence the classification task is defined as the prediction on the subgraphs. Another important distinction is that the superpixel images require to construct the connectivity between chosen nodes. 
 A comparison between the grid and superpixel images is shown in Fig. \ref{fig:grid_super}, where 75 superpixels are processed as the representation of the image.

One of the important steps when processing the superpixel data is learning effective graph embedding extracted from the graph. To demonstrate the ability of our model when predicting based on the superpixel images, we empirically follow the convolutional approach utilized in SplineCNN \cite{DBLP:conf/cvpr/FeyLWM18} to further aggregate  the connectivity of superpixels. The trainable kernel function based on B-splines can make the most use of the local information in the graph and filter the input into a representative embedding. Similar to the framework proposed in Fig \ref{fig:extended}, the experiments on superpixel images also follow the structures as grid image experiments, where the convolutional layers \& IF neurons enable the spike representations, and the fully connected layers \& LIF neurons are responsible for the classification results.

\begin{figure}[p]
\centering
\subfigure[Outputs of Bernoulli encoder in grid images]{
\begin{minipage}[p]{1\linewidth}
\centering
\includegraphics[width=3.2in]{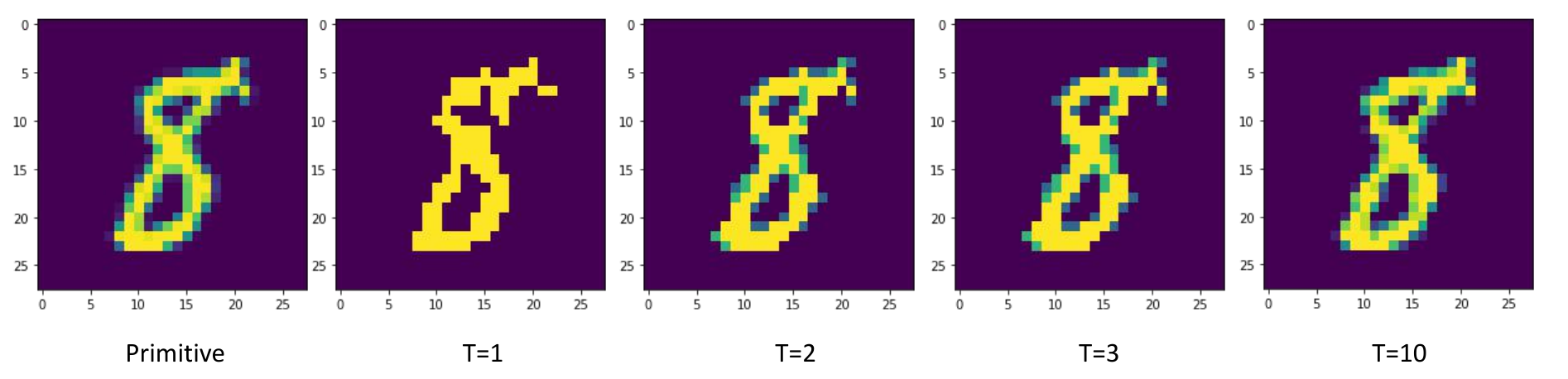}
%\caption{fig1}
\end{minipage}%
}%
\quad
\subfigure[Outputs of trainable encoder in grid images]{
\begin{minipage}[p]{1\linewidth}
\centering
\includegraphics[width=3.2in]{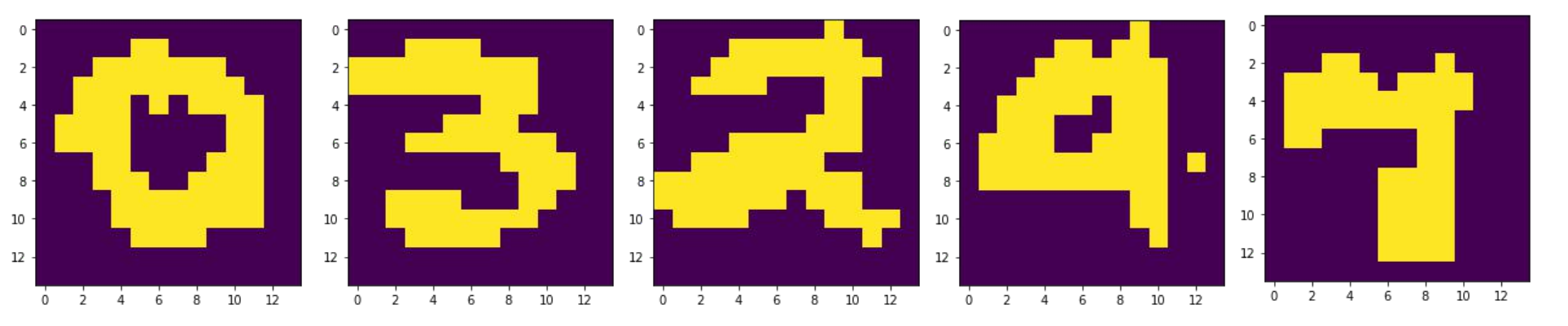}
%\caption{fig2}
\end{minipage}%
}%
\quad
\subfigure[Outputs of trainable encoder in superpixel images]{
\begin{minipage}[p]{1\linewidth}
\centering
\includegraphics[width=3.2in]{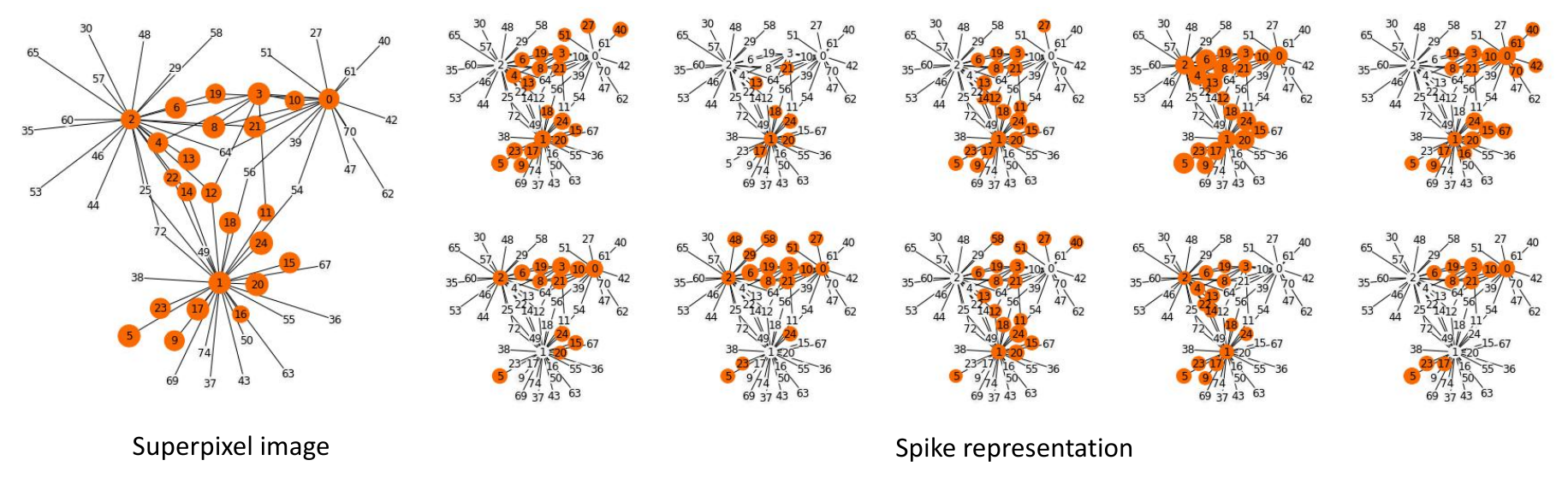}
%\caption{fig2}
\end{minipage}
}%
\centering
\caption{Visualization of the spike trains generated by the spike encoder. We extract these features from the MNIST dataset for demonstration. Grid images:
(a) shows the spike trains from a simple Bernoulli encoder, and we list the different time steps which indicate different precision. (b) depicts the spikes from the trainable spike encoder, in which the overall shape patterns are learned. Superpixel images: (c) demonstrates the spikes from the trainable encoder, and the encoding results indicate the successful detection of local aggregation. }
\label{fig:encoding result}
\end{figure}

In addition, we also provide a unique perspective to understand the mechanism of our model. In particular, our spike encoder can be regarded as a sampling process using spike train representation. The scenario of the image graph provides us an ideal chance to visualize the data processing in our model. Regarding the experiments of grid and superpixel images, we extract the outputs of our spike encoder and visualize them in Fig. \ref{fig:encoding result} (c), along with other observations. First, the Bernoulli encoder mentioned above can be viewed as a sampling process with respect to the pixel values. As the time step increases, the encoder almost rebuilds the original input. However, the static spike encoder can not capture more useful features from the input data. Thus, our trainable encoder performs the convolution procedure and stimulates the IF neurons to fire a spike. As shown in  Fig. \ref{fig:encoding result} (b) and (c), by learning the convolutional parameters in the encoder, the spike encoder successfully detects the structure patterns and represents them in a discrete format.

\paragraph{Spike encoder for recommender systems.}\label{recommender system} 
Much research has tried to leverage the graph-based methods in analyzing social networks \cite{DBLP:journals/corr/BergKW17,DBLP:conf/sigir/Wang0WFC19,DBLP:conf/sigir/0001DWLZ020}.
To this end, we extend our framework to the recommender systems, where users and items form a bipartite interaction graph for message passing. 
%In order to predict the preference from certain users, completing the interaction matrix can be a significant task. In this work, 
We tackle the rating prediction in recommender systems as a link classification problem. %Many recent works use the deep learning methods to reconstruct the rating links between users and items, which inspires us to apply our SNN-based framework to conduct the link prediction. 
Starting with MovieLens 100K datasets, we take the rating pairs between users and items as the input, transform them into suitable spike representations, and finally output the classification class via firing rate. To effectively model this graph-structured data, we build our trainable spike encoder based on the convolutional method used in GC-MC \cite{DBLP:journals/corr/BergKW17}. In particular, GC-MC applies a simple but effective convolutional approach based on differentiable message passing on the bipartite interaction graph, and reconstruct the link utilizing a bilinear decoder. %This network topology, which has the potential to act in cooperation with SpikingGCN, motivates us to complete the rating matrix in an SNN manner.

%\section{Source code} \label{code}
%Our source code is available at \ulr{https://anonymous.4open.science/r/SpikingGCN-1527}.

% \fi
\end{document}